\documentclass[utf8]{FrontiersinHarvard} 
\makeatletter
\@ifundefined{verse}{%
}{}
\makeatother
\usepackage{url,hyperref,lineno,microtype}
\usepackage[onehalfspacing]{setspace}
\usepackage{lipsum}
\usepackage{multirow}
\usepackage{booktabs}

\usepackage{titlesec}
\titleformat{\paragraph}
  {\normalfont\normalsize\bfseries}
  {\theparagraph}
  {1em}
  {}

\titlespacing*{\paragraph}
  {0pt}{3.25ex plus 1ex minus .2ex}{1.5ex plus .2ex}

\usepackage{colortbl}
\usepackage{makecell}
\usepackage{graphicx}

\usepackage[utf8]{inputenc}
\usepackage{amssymb}
\DeclareUnicodeCharacter{25A2}{$\square$}
\usepackage{color}
\usepackage{dialogue}

\usepackage{array}
\usepackage{comment}
\usepackage{tcolorbox}
\newcounter{promptno}[section]
\newlength\mystoreparindent
\newenvironment{prompt}[1][]
{
  \setlength{\mystoreparindent}{\the\parindent}
  \setlength{\parindent}{0pt}
  \refstepcounter{promptno}
  \par\medskip
  \noindent
  \begin{tcolorbox}[left=1pt,right=1pt]
  \textsc{{template \small\thesubsection.\thepromptno}}\\
  \small
  \tt
}{
  \end{tcolorbox}
  \setlength{\parindent}{\mystoreparindent}
  \medskip
}

\def\keyFont{\fontsize{8}{11}\helveticabold }
\def\firstAuthorLast{Kranti {et~al.}} 
\def\Authors{Chalamalasetti Kranti\,$^{1,*}$, Sherzod Hakimov\,$^{1}$ and David Schlangen\,$^{1,2}$}
\def\Address{$^{1}$Computational Linguistics, Department of Linguistics, University of Potsdam, Germany \\
$^{2}$German Research Center for Artificial Intelligence (DFKI), Berlin, Germany}
\def\corrAuthor{Chalamalasetti Kranti}

\def\corrEmail{kranti.chalamalasetti@uni-potsdam.de}

\begin{document}
\onecolumn
\firstpage{1}

\title{Teach, Abstract, Apply: A Framework for Dialogue-Driven Task Learning with Collaborative Robots} 

\author[\firstAuthorLast ]{\Authors} 
\address{} 
\correspondance{} 

\extraAuth{}

\maketitle

\begin{abstract}
Collaborative robots (cobots) are designed to be safer and more interactive than traditional industrial robots, share the same workspace as humans, and assist them in a range of shop-floor tasks such as palletization, joint assembly, and part inspection. Many of these tasks involve repetitive workflows that may evolve with changes in production environments and task specifications. These requirements motivate the need for intuitive interfaces that allow workers to modify workflows on the shop floor and generalise these routines for future use. This work introduces the Teach-Abstract-Apply (TAA) framework, which enables human operators to (a) teach new workflows to cobots in natural language, (b) abstract the taught workflows into skills for later use, and (c) apply these skills in different settings. As a case study, we apply the TAA framework to a collaborative structure-building task and evaluate it with both open-weight (\texttt{Qwen3-Coder}) and closed (\texttt{GPT5.2-Codex}) models in self-play and cross-model configurations. Our results show that skill abstraction improves task success rates and reduces interaction cost. Beyond automated evaluation, a preliminary human-in-the-loop study suggests that the pipeline also works with human instructions.

\tiny
 \keyFont{ \section{Keywords:} collaborative dialogue systems, human-robot interaction, natural language dialogue, skill learning, skill reuse, program synthesis}
\end{abstract}

\section{Introduction}
\label{sec:introduction}
Industry 4.0~\citep{russmann2015industry, xu2018industry} suggests a transformation from conventional automation to the use of smart machines and smart factories in manufacturing. Its vision includes collaborative robots (cobots) assisting human coworkers~\citep{colgate1996cobots} on factory floors to enhance productivity, flexibility and efficiency. Unlike traditional industrial robots, these cobots are designed with enhanced safety mechanisms~\citep{606888, DBLP:conf/icra/AkellaPCWNWHPP99} and simplified programming requirements~\citep{981926, 1045667} that enable them to work alongside human coworkers in the same workspace and reduce the need for expert robot programmers in operating them. However, real-world human-robot interaction is still far from achieving such seamless collaboration~\citep{9493209, segura2025work}. Various research approaches~\citep{7745145, 10.1007/978-3-030-28619-4_26, 9889345, DBLP:conf/hri/WhiteM24} have therefore focused on lowering the barriers to using cobot interfaces. With many industrial robots deployed on assembly floors, it is important to enable effective collaboration~\citep{AALTONEN20191151, 10.1145/3610977.3634937} between humans and robots.

Recently, human-robot collaboration workflows have been improved by advancements in large language models (LLMs)~\citep{ZHANG2023100131, DBLP:conf/hri/KimLM24}, with a focus on immediate, specific actions for \emph{``here and now''} scenarios~\citep{DBLP:conf/icml/DriessXSLCIWTVY23}. These workflows usually operate by taking commands from users~\citep{DBLP:conf/corl/IchterBCFHHHIIJ22}, with the backend LLM analysing and reasoning about them~\citep{DBLP:conf/icra/LiangHXXHIFZ23}, preparing a step-by-step plan~\citep{DBLP:conf/icml/HuangAPM22, DBLP:journals/tmlr/WangX0MXZFA24} that considers the current environment, and generating a program~\citep{DBLP:conf/icra/SinghBMGXTFTG23} to perform the task. However, these methods focus primarily on executing the current task rather than retaining and reusing these behaviors, which is the core intent of teaching custom behaviors to the robot. Furthermore, in industrial environments, \textbf{it is useful to teach a task as well as to capture the behavior so it can be re-deployed in new settings with minimal user intervention}~\citep{bauer2016lightweight, VILLANI2018248, robots2024cobots, 10.1145/3610977.3634937}. Without a mechanism to generalize and reuse learned behaviors, robots must be re-taught or re-programmed each time the environment changes, limiting scalability.

Teaching is often iterative, involving back-and-forth clarifications~\citep{DBLP:conf/aaai/PadmakumarTSLNG22} and corrections that lead to incremental refinement~\citep{chiyah-garcia-etal-2024-repairs}. Prior work on conversational programming~\citep{DBLP:journals/ijst/Desilets01, DBLP:journals/corr/abs-1910-11471} addresses parts of this process by enabling programs to be iteratively specified and refined through dialogue~\citep{cypher1993watch}. Building on this, we propose the \textit{``TAA: Teach, Abstract, Apply''} framework that enables workers to talk to a robot in natural language~\citep{DBLP:journals/ras/LauriaBKK02, GOROSTIZA20111102}, iteratively refine the workflow~\citep{10.1109/ICRA.2019.8794287}, and thereby teach new behaviors through a multi-turn dialogue~\citep{DBLP:conf/vl/GeDSLHS24}. As the name suggests, the TAA framework consists of three stages: Teach, Abstract, and Apply. \textit{Teach} refers to the process of a worker engaging in a multi-turn interaction with the robot, with the robot extracting relevant information from the interaction. \textit{Abstract} refers to the process of capturing and generalising the extracted information into a behavior that can be applied across different environments. \textit{Apply} refers to the process of applying these learned behaviors in different settings with minimal user intervention.

\begin{figure*}[h]
\centering
\caption{Overview of the Teach and Abstract steps represented through the structure building task in a simulated environment. Teach: the worker uses natural-language to teach the Cobot an assembly arrangement;
Abstract: successful interaction traces are abstracted into a reusable skill.
}

\label{fig:cla_cl_overview}
\includegraphics[width=\textwidth]{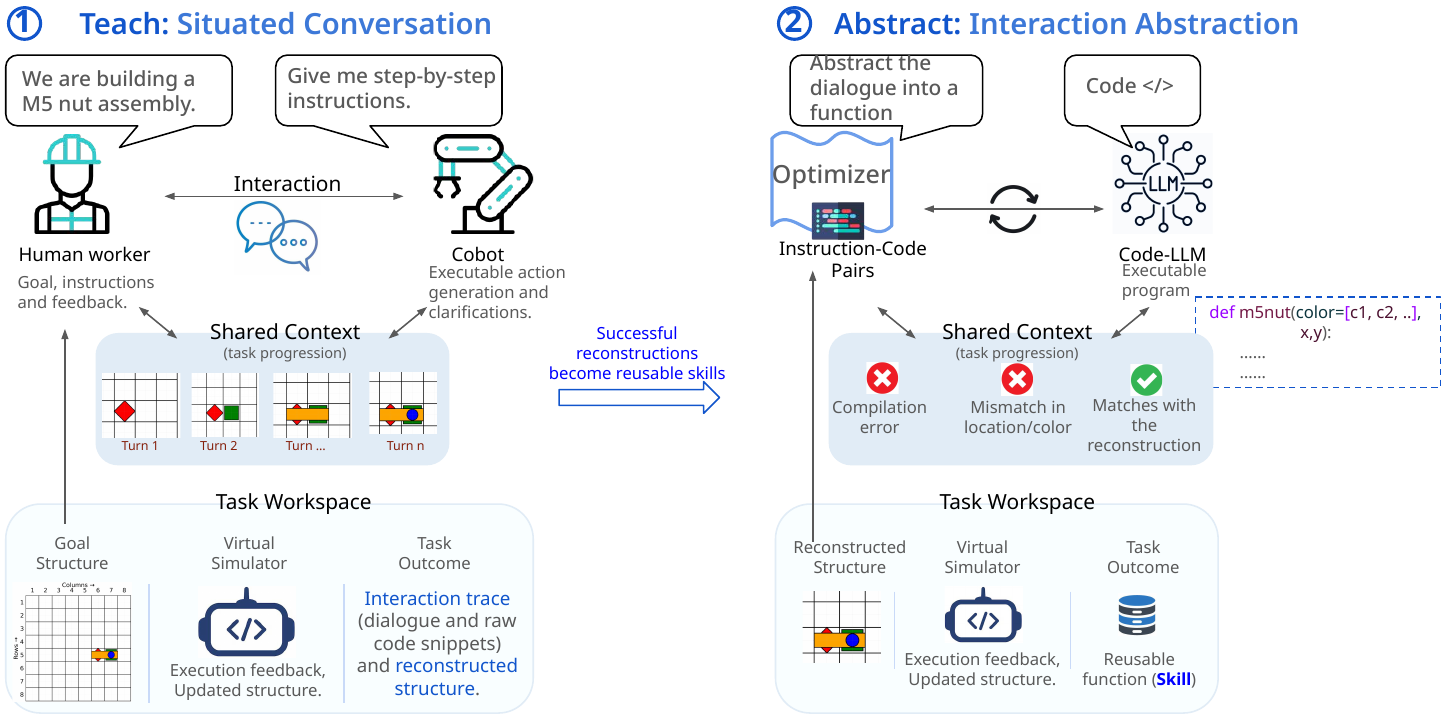}
\end{figure*}

\begin{figure*}[h]
\centering
\caption{Overview of the Apply step represented through the structure-building task in a simulated environment. Reuse: refers to using an existing skill at a new location; Repeat: refers to using an existing skill with new properties — the washer color is set to `green' and the skill is repeated across a set of locations in a pattern.}
\label{fig:cla_a_overview}
\includegraphics[width=\textwidth]{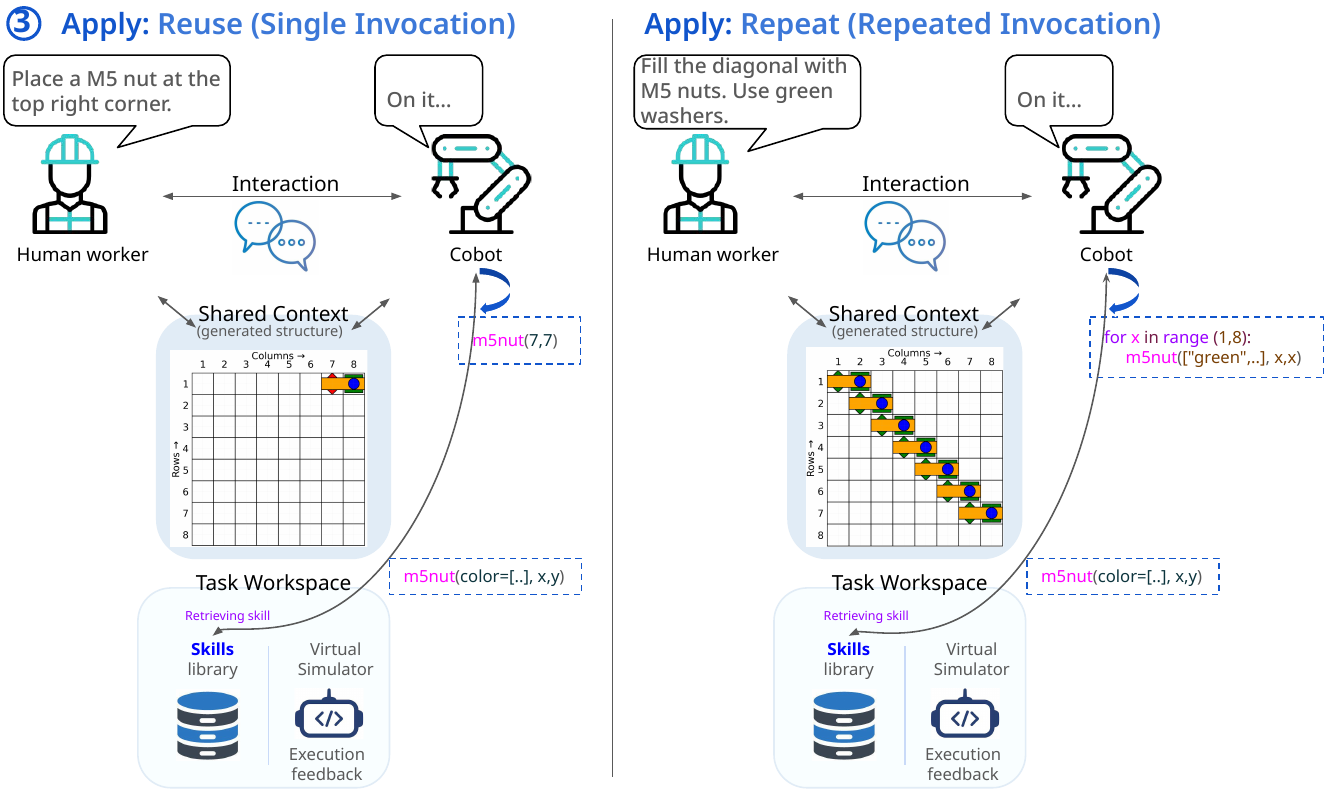}
\end{figure*}

We test the TAA framework in a controlled setting through a collaborative structure-building task (see Figure~\ref{fig:cla_cl_overview}) inspired by real-world industrial assembly scenarios, in which a worker converses with the robot to build a structure in a grid-based environment. These structures are composed of components such as washers, nuts, screws, and bridges. During the Teach step, the worker communicates the structure to be built in natural language in a step-by-step fashion, and the robot semantically translates these instructions into executable code. The outcome of the Teach step is an interaction trace consisting of a set of instruction-action pairs, which are then used by the Abstract step to capture and generalise the code into a persistent function that can be adapted to different environments. The Apply step focuses on evaluating how well the robot can reuse these persistent functions in different settings without requiring the worker to teach it again from scratch. To isolate the effectiveness of this framework, the task is evaluated in a controlled environment without a physical robot. An instruction-tuned Code LLM generates executable code snippets for robot manipulation, which are executed in a virtual simulator, thereby for now avoiding the perception and control issues of a real robot.

With this controlled task setting, we investigate the following research questions: (R1) Can LLM-based interpretation together with a dialogue management framework realise acquisition of executable behaviours through multi-turn natural-language dialogue? (R2) Can LLM-based interpretation be used to abstract from the concrete teaching situation a function that generalises to new spatial contexts and reused without re-specification? (R3) Can learned behaviors improve task success and reduce interaction cost compared to teaching from scratch? Our contributions are threefold: (1) we present TAA, a dialogue-based framework for teaching new behaviors grounded in executable programs; (2) we introduce a behavior abstraction mechanism that generalises demonstrated behaviors into reusable functions; and (3) we provide a comprehensive evaluation, including reconstruction, reuse, and repeat experiments, along with a pilot human study, demonstrating that learned behaviors enable persistent reuse, reduce interaction cost, and improve task performance compared to context-only conversational programming.

\section{Related Work}
\label{sec:relatedwork}

To make cobots more accessible and make full use of their collaborative potential~\citep{tsarouchi2016human, ELZAATARI2019162, ZHANG2023100131}, prior work has introduced a variety of interfaces, including text-based~\citep{10.1145/3610977.3634969}, voice-based~\citep{ionescu2023programming}, visual~\citep{9320477, 9889345}, multi-modal~\citep{beschi2019capirci, FOGLI2022102234, 11534298}, and kinesthetic teaching~\citep{AJAYKUMAR2021103845} interfaces. However, these interfaces primarily support interaction for the current task. Our work extends such interactions to generate persistent representations that can be reused in later tasks.

\subsection{Collaborative Dialogue Systems}
Human-robot collaboration systems have evolved from instruction-only style~\citep{DBLP:journals/ras/LauriaBKK02} to interactive style~\citep{10.1145/1180995.1181009,GOROSTIZA20111102} that showcases the necessity of back-and-forth negotiation. \citet{https://doi.org/10.1609/aimag.v37i4.2684, 10.5555/3304415.3304417} discuss the inherently ambiguous nature of human instructions, which motivates the need for correction and clarification as core interaction patterns~\citep{10.1109/ICRA.2019.8794287, kane-etal-2022-system, chiyah-garcia-etal-2024-repairs}, the robot must actively seek and incorporate feedback across turns. More recently~\citet{10.1007/s10579-024-09784-2} explores establishing common ground across different modalities of information in human-robot dialogue. Although these systems advance collaborative interaction, the knowledge learned through dialogue is not retained as an executable representation. Our work builds on these systems by extracting and persisting learned knowledge as reusable executable behaviors.

\subsection{Conversational Programming}
Prior work~\citep{5391077, LEGGETT1984493, DBLP:journals/ijst/Desilets01} proposed using different modalities for translating natural language instructions into programs. With recent advancements in LLMs, program synthesis has become widely explored~\citep{DBLP:journals/corr/abs-2107-03374, DBLP:journals/corr/abs-2108-07732, DBLP:journals/tosem/HouZLYWLLLGW24}. \citet{DBLP:conf/iclr/NijkampPHTWZSX23} explore multi-turn interactions for improving overall program synthesis quality. Several works~\citep{DBLP:conf/icml/HuangAPM22, DBLP:conf/corl/IchterBCFHHHIIJ22, DBLP:conf/icra/LiangHXXHIFZ23, DBLP:conf/icra/SinghBMGXTFTG23, DBLP:conf/irc/TobergC23,DBLP:conf/icra/MacalusoCC24, 10.1145/3757279.3785550} extend these ideas to generate executable robot tasks and plans from natural language instructions. However, these works focus primarily on context-bound execution and often assume users with programming knowledge. Our work instead targets factory workers, who may lack programming background but have domain expertise~\citep{AALTONEN20191151, bauer2016lightweight, VILLANI2018248, 10.1145/3610977.3634937}. Furthermore, the outcome of the dialogue is captured as a persistent, reusable skill representation in the form of an executable program.

\subsection{Reusable Skills}
Once deployed in the real world, collaborative robot routines evolve over time, which traditionally involves re-programming~\citep{1456949,biggs2003survey,VILLANI2018248,10.1145/3466819,robotics11010009}. To reduce this burden, \citet{DBLP:journals/ijrr/KonidarisKGB12} proposed segmenting demonstrations into reusable skill trees that allow users to build new sets of skills. \citet{DBLP:journals/ijrr/LioutikovNMP17} learn a library of movement primitives using probabilistic methods. Although these approaches aim to discover reusable primitives from demonstrations, they still depend on physical robot access and predefined model structures and representations. Leveraging recent LLM progress, several works use them to generate robot programs~\citep{pmlr-v235-mu24a, DBLP:journals/corr/abs-2405-20179}. \citet{DBLP:journals/tmlr/WangX0MXZFA24} demonstrate building a skill library that allows reusing learned skills; however, the stored skills are specialized rather than generalized. \citet{DBLP:conf/icra/Tziafas024} extend this to generalizable skills in lifelong learning. Moreover, recent advances in coding agents have introduced user-editable skills in text files with recipes for task development. This has led to the evolution of an open-standard (\texttt{SKILL.md}~\footnote{https://github.com/agentskills/agentskills}) file that describes a skill and its associated tools and workflows. Similar ideas have been explored in robotics, to generate, refine and reuse robotic behaviors~\citep{DBLP:journals/corr/abs-2607-00272, hu2026simex, pmlr-v306-fu26u}. While TAA 
shares the idea of maintaining persistent skills with these works, it differs in how these skills are systematically evaluated across different application settings through reuse and repeat tasks. This is a necessary assessment because the goal is not only to acquire a skill through teaching, but also to redeploy it in new settings.

Taken together, these lines of research establish the building blocks for TAA. Collaborative dialogue systems demonstrate the usefulness of multi-turn interaction for clarifying and grounding user intent, but they do not retain what is learned. Conversational programming shows how natural-language interactions can be used to iteratively specify and refine executable programs, but primarily focuses on generating programs for the current task. Works on skill reuse demonstrate the value of persistent, reusable robot behaviors, but does not evaluate how these skills transfer under systematic task modifications. TAA brings them together by enabling users to teach robots new behaviors through multi-turn dialogue, abstracting the interaction into persistent executable skills, and evaluating their reuse across systematic task modifications.

\section{Methodology}
\label{sec:methodology}

\subsection{TAA: Teach, Abstract, Apply}
Building on human–robot collaborative dialogue systems~\citep{10.1145/1180995.1181009, GOROSTIZA20111102,robotics11010009,kane-etal-2022-system,10.1007/s10579-024-09784-2}, we formulate the Teach-Abstract-Apply (TAA) framework as three distinct and sequential steps: Teach, Abstract, and Apply, where the outcome of each preceding step provides input to the next step.

The \textbf{Teach} step begins with a goal that is known only to the worker and may be represented in different forms, such as an image (e.g., a schematic diagram), text (e.g., a design document), voice note (e.g., a description of the required functionality), or a combination of one or more of these modalities. The worker interacts with the robot in natural language and incrementally guides the robot towards the goal. The interactions usually span multiple turns and include clarifications, corrections and acknowledgments from both the worker and the robot. We frame this interaction as a programming process which is refined iteratively. The outcome of the Teach step is an \textit{interaction trace} that encodes the workflow. This is followed by the \textbf{Abstract} step, which generates a persistent representation of the workflow. Necessary information from the interaction trace is extracted and generalised so that the retained representation can be extended to different environments. The outcome of the Abstract step is a \textit{skill} (e.g., a program, a \texttt{.md} file etc.) that is available for later use. And the \textbf{Apply} step follows, which applies the skills from the previous step in new settings. The robot is equipped with these new skills and the evaluation  focuses on whether the skills can be adapted and applied to the new settings (e.g., changing the attributes in a workflow, repeating the workflow in different patterns, or building a new workflow using the current workflow etc.) without teaching from scratch. 

\subsection{Structure Building Task}
\citet{kranti2024towards} proposed a structure-building task (SARTCo) in which an instruction giver (\textit{Programmer}) collaborates with an instruction follower (\textit{Cobot}) to reconstruct a goal structure, which consists of different industrial-assembly-style configurations. The Programmer receives the goal structure, represented on a \(8 \times 8\), \(2.5D\)~\footnote{A 2D grid with vertical stacking.} grid, and instructs the Cobot in natural language on how to reconstruct it. The grid is represented as a Python array, and each grid cell can be accessed using its row and column values, \textit{e.g.,} [1,2] or [3,4]. The Cobot translates the instructions into code snippets, which, when executed in a simulator, generate the resulting structure on the grid. They explore generating first-order (primitive code) and higher-order (abstract programs) programs from single-turn instructions. We use this dataset for our experiments.

\subsection{Realising TAA Framework}
\label{subsec:methodologytaa}
Our objective is to enable workers to teach robots persistent workflow representations through natural language using TAA. For this purpose, we instantiate TAA through the structure building task in a controlled setting. We use instruction-tuned Code LLMs as both the Programmer and the Cobot. The LLM powering the Programmer receives the goal structure in textual form and instructs the Cobot in natural language. The Cobot receives Programmer's instruction along with the current state of the reconstructed board in textual form and generates code snippets. These code snippets are executed in the simulator, and the resulting structure is returned to the Programmer in textual form as a follow-up response. We use the \texttt{clem:todd} framework~\citep{kranti-etal-2025-clem} to facilitate these interactions which follow a specific JSON schema (\textit{e.g., code, clarification, acknowledgment}). At the end of the interaction, the resulting instruction-code pairs are used to generate a higher-order Python function that is saved as a skill. These saved skills are evaluated for their extendability to new tasks. As mentioned in Section~\ref{sec:introduction}, we do not use a physical robot in this controlled evaluation.

\subsubsection{Teach: Teaching through Dialogue}
\label{subsubsec:reconstruction}
\begin{figure*}[h]
\centering
\caption{An example interaction trace showing how the conversation proceeds. At Turn 1, the Programmer instructs the Cobot to place a red screw on top of a yellow washer at position [1,3]. However, the Cobot responds with an action that places only a washer and also uses the wrong color. The Programmer corrects the color and proceeds with the next step. At Turn 3, the Cobot asks for clarification about the screw color, although it was already specified in Turn 1. In the next turn, the Programmer repeats the details. The shared context shows how the reconstructed board is updated after each Cobot action. When the Programmer responds with \texttt{DONE}, the interaction is completed.}
\label{fig:interactiontrace}
\vspace{6pt}
\includegraphics[width=\textwidth]{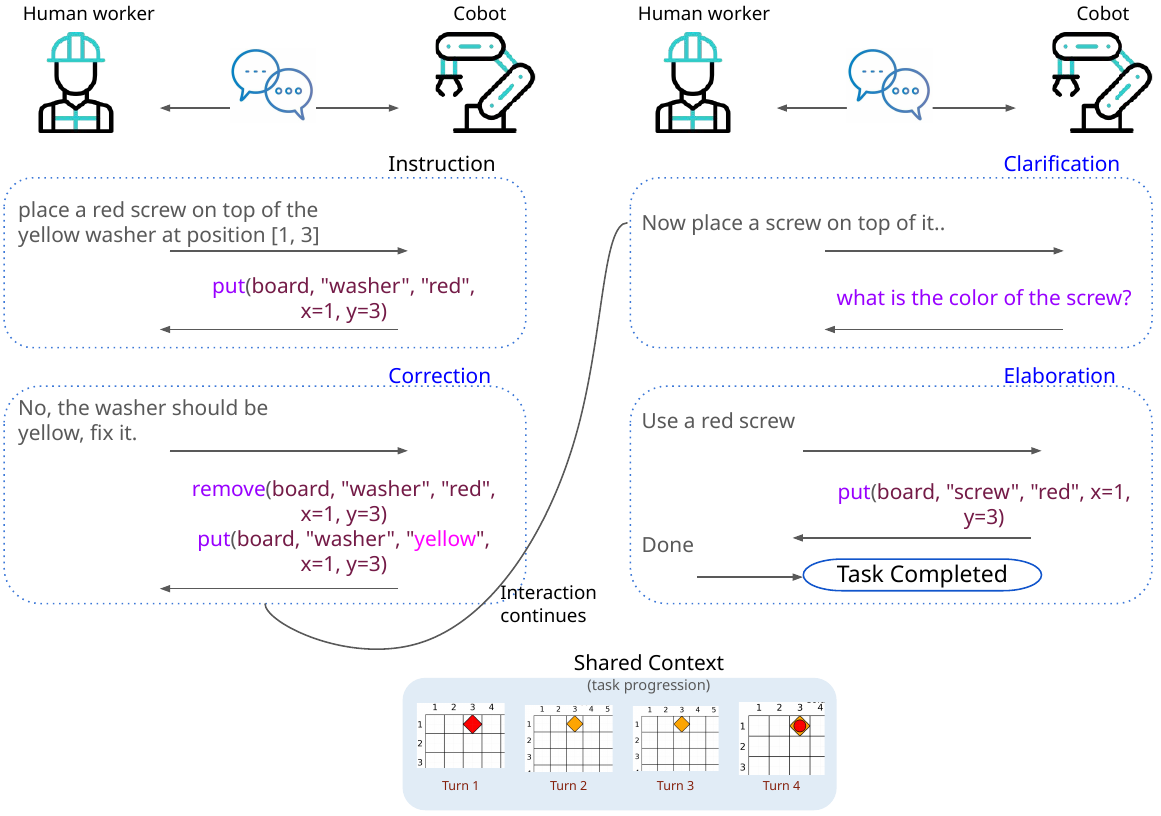}
\end{figure*}

A conversation demonstration (see Figure~\ref{fig:interactiontrace}) refers to an incremental teaching process through which the worker teaches the cobot new manipulation routines for building a given goal structure (also referred as a \textit{board}). The Programmer may be a human, LLM, or rule-based generator, while the Cobot may be a human or LLM, are treated as the two players in the structure building task, and the Game Master (GM) coordinates the turn-level interaction between them. The task concludes when Programmer responds with \texttt{DONE}, or when the maximum turn limit ($15$) is reached.

\paragraph{Programmer: Instruction Giver}
In line with existing work~\citep{chalamalasetti-etal-2023-clembench, DBLP:conf/slt/KaziLZHT24}, we use LLMs to simulate instructors. We conducted multiple ablation studies to design the final prompt structure (details in Section~\ref{subsec:ablationstudies}). Programmer is prompted~\footnote{Refer to the Supplementary Material for detailed prompts.} in a zero-shot setting with information about the environment, the task constraints, and the goal board to reconstruct. The goal board is represented in textual format for a text-based LLM and describes the stacking order and cell-level details, including the occupied elements and their colors. Based on the ablation studies, the prompt included a \textit{difference board}, that shows the difference between the goal board and the current reconstruction board for each cell and stacking level. In the first turn, the difference board is \texttt{None}; in subsequent turns, it is updated with the relevant differences.

When instructions are ambiguous or incomplete, Cobot requests clarification, and Programmer elaborates or corrects the description. If Cobot makes an error, such as placing unintended shapes in the grid, colors, or spatial positions, Programmer provides corrective feedback so that subsequent outputs align with the intended goal. Through this iterative process, Programmer and Cobot collaboratively reconstruct the goal board.

\paragraph{Cobot: Instruction Follower}
Cobot translates user instructions into executable code. We use instruction-tuned code LLMs for this purpose. Cobot is prompted with information about the environment, the task constraints, and the current state of the reconstruction board. Similar to Programmer, the current reconstruction board is represented in textual format, describing the stacking order and cell-level details, including the occupied elements and their colors.

For a given input instruction, Cobot is expected to either produce the corresponding action sequence (\textit{e.g.,}, input: ``place a red washer in cell 4,3''; output: \texttt{put(board, `washer', `red', x=4, y=3)}) or ask for clarification (\textit{e.g.,}, ``there is already a nut present in the same location. do I need to remove the nut before placing a washer or stack on it?''). The generated action sequences are executed in a virtual simulator. If execution fails, Cobot receives feedback about the error and either corrects the code or asks for clarification when the error cannot be resolved without additional context from Programmer. If execution continues to fail after three attempts, the interaction is terminated and the test instance is treated as a failure. Cobot can also acknowledge Programmer queries (\textit{e.g.,}, input: ``do you know how to place a washer?''; output: ``Yes, I know!'').

\paragraph{Game Master: Interaction Coordinator}
The Game Master (GM) manages the turn-level interaction between Programmer and Cobot. At the beginning of each turn, GM provides Programmer with the goal board and the difference board. Programmer then gives a natural language instruction to guide the reconstruction, and Cobot responds by translating the instruction into executable Python code, asking for clarification, or providing an acknowledgment when appropriate. GM validates each player response to ensure that it follows the required format; if a response does not satisfy the format requirements, the interaction is aborted. GM also executes Cobot's generated code in the virtual simulator and shares the updated reconstruction board with both players in the subsequent turn. When execution fails, GM forwards the error to Cobot, which can revise the code or ask Programmer for clarification. The interaction continues until Programmer responds with \texttt{DONE}, the maximum turn limit ($15$) is reached, or execution fails after three attempts. Once Programmer responds with \texttt{DONE}, GM evaluates task success by comparing the reconstructed board with the goal board at each cell location, including component color and stacking order. If the reconstruction is successful, GM forwards the instruction-code pairs, consisting of each Programmer instruction and its corresponding Cobot-generated action, to the Optimizer module for further processing. The prompt structure, number of execution retries, and overall task interaction are based on the ablation studies, discussed in Section~\ref{subsec:ablationstudies}.

\subsubsection{Abstract (Optimizer): Generalisable Function Generator}
\label{subsec:optimization}
\begin{figure*}[h]
\centering
\caption{Overview of reusable function generation. Given the instruction-response pairs and a predefined function signature, the Code LLM abstracts the primitive code snippets into a reusable higher-order function. The generated function is then evaluated by comparing the structure it reconstructs with the structure produced by the input code snippets, evaluating                                                                                each cell for component type, color, and stacking level.}
\label{fig:optimization}
\vspace{6pt}
\includegraphics[width=\textwidth]{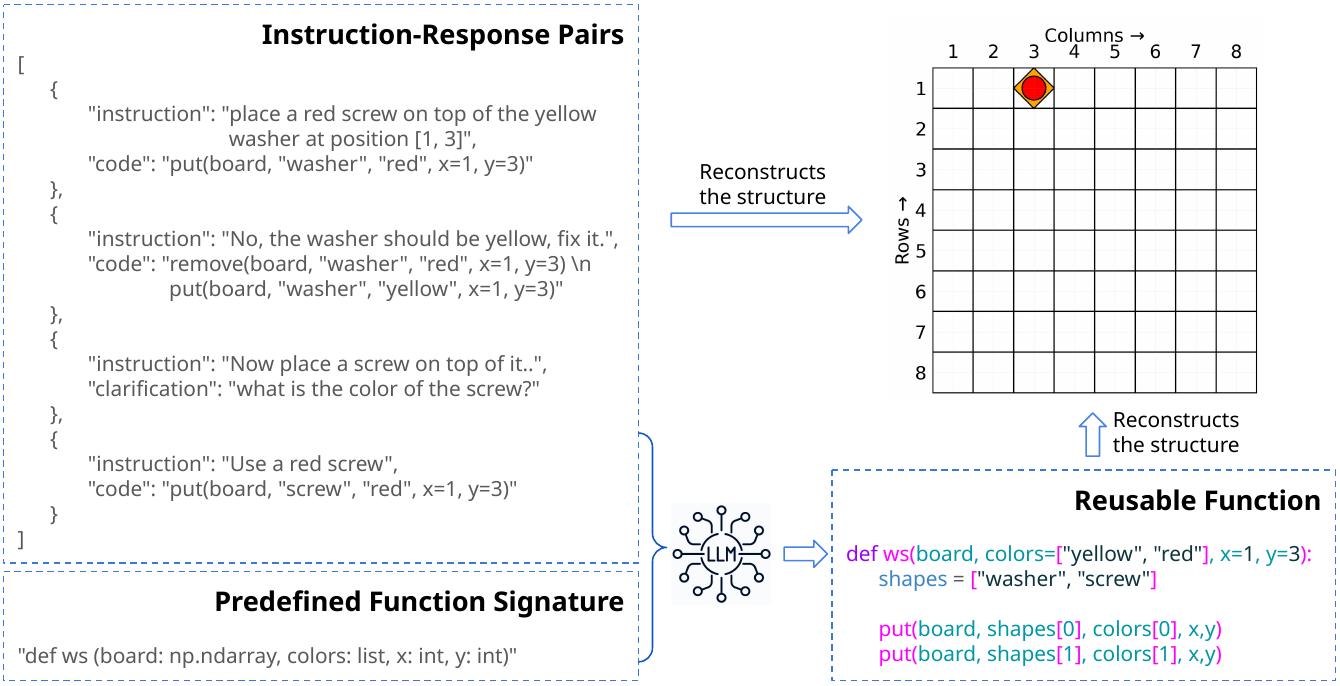}
\end{figure*}

Only successful reconstruction interactions are used for optimization, since they correspond to cases where the goal board was reconstructed correctly. Unsuccessful or incorrect interactions are excluded. Optimizer receives the sequence of instruction-response pairs collected during reconstruction, where each Programmer instruction is paired with the corresponding Cobot-generated action as shown in Figure~\ref{fig:optimization}.

The Cobot output contains primitive placement sequences and may also reflect clarifications or corrections made during the interaction. Although these sequences are sufficient to reconstruct the goal board, they cannot be reused in later tasks unless they are converted into a reusable representation. We therefore abstract recurring action patterns into higher-order Python functions that produce the same goal board state as the primitive code sequences.

Optimizing code into reusable abstractions is often time-consuming and requires human expertise. Prior work~\citep{DBLP:journals/corr/abs-2309-03409, 11091799, 11029992} has shown that LLMs can support optimization processes. Building on this, we follow a similar LLM-as-optimizer paradigm and use an instruction-tuned code LLM to iteratively transform the generated code into a reusable skill function. The optimizer is queried in a zero-shot setting with a prompt that includes information about the environment, task constraints, and instruction-response pairs from the reconstruction task. The output is a Python function, referred to as a skill, that encapsulates the manipulation routine in a reusable form for subsequent tasks.

The optimizer receives the full sequence of instruction-response pairs without pre-processing (see Figure~\ref{fig:optimization}). Since the sequence may include instructions that correct or clarify earlier ones, the optimizer is prompted to read the entire interaction, identify which code reflects the final intended actions, and discard any code that was later replaced or fixed. For example, an instruction may place a washer using \texttt{put(board, `washer', `red', x=1, y=3)}. If a later instruction corrects the washer color to yellow with \texttt{put(board, `washer', `yellow', x=1, y=3)}. The final code used for optimization would therefore include only \texttt{put(board, `washer', `yellow', x=1, y=3)}. The prompt also instructs the optimizer to follow a predefined function signature. For illustration, a generated skill may be defined as \texttt{def <object\_name>(board, colors, x, y)} and be used as \texttt{ws(board, [`yellow', `red'], 1, 3)}. The generated function can then be stored and reused in subsequent tasks (Sections~\ref{subsubsubsec:reuse} and \ref{subsubsubsec:repeat}).

For a given input, Optimizer is expected to generate a Python function that produces the same board state as the input primitive code sequences when executed in the virtual simulator. Equivalence is determined by comparing the final board states produced by the generated function and the primitive code sequences. If the Optimizer output fails to execute, it receives feedback about the error and can revise the code. If it cannot correct the error within three attempts, the task is aborted and the test instance is treated as a failure. The task is considered successful when the Optimizer output replicates the board state and is considered failed when the maximum turn limit ($10$) is reached.

\subsubsection{Apply}
\label{subsubsec:methodologyapply}
\paragraph{Reuse: Systematic Evaluation of Skill Generalisability}
\label{subsubsubsec:reuse}
Once cobots are deployed in the real world, onsite requirements may differ from their pre-programmed routines due to customization needs or changing task constraints. These changes can include a new product variant, a change in assembly geometry, such as a V-shape instead of a T-shape, or an added inspection step, such as checking alignment before fastening. Adapting to such changes often requires a programming expert~\citep{10.1145/3466819, robotics11010009} or the use of complex interfaces~\citep{DBLP:journals/ei/AkkaladeviPPIH19, 9800018, george2023task, DBLP:conf/hri/AyalewWLUS25}, which can be costly~\citep{barravecchia2023general} and slow~\citep{10.1145/3466819}.

We design the reuse task to study whether skills generated by the Optimizer can be adapted under controlled modifications. Specifically, we evaluate changes in color and spatial information, reflecting incremental adjustments common in assembly workflows. We are interested in: (a) can the model identify the required skill from the list of available skills? and (b) can the model use the selected skill with the correct modifications for the new requirement?

Similar to reconstruction, the reuse task consists of two players, Programmer and Cobot, interacting over a multi-turn dialogue to replicate a goal board. The skills generated by the Optimizer are made available in Cobot's context. Each skill is represented by its function signature and a single-line description of the object it builds. Programmer is prompted to ask whether Cobot is aware of the relevant skill before guiding it to build the modified object using the available skills. Cobot is expected to identify the relevant skill from its context and generate a function call with the correct arguments. This function call is executed in the virtual simulator using the corresponding skill definition learned during optimization. If the corresponding skill is not available in the library, we use the ground truth function definition to execute the function call.

If Cobot cannot identify a suitable skill, Programmer provides step-by-step instructions, and the interaction falls back to the Teach reconstruction loop. The task is completed when Programmer signals completion or when the maximum turn limit ($15$) is reached. Once Programmer responds with \texttt{DONE}, GM evaluates task success by comparing the reconstructed board with the goal board at each cell location, including component color and stacking order.

\paragraph{Repeat: Systematic Evaluation of Repeated Skill Use}
\label{subsubsubsec:repeat}
As robots are often used for repetitive tasks, we extend our study to examine how learned skills generalize to goal boards with repeated spatial patterns. The repeat task consists of goal boards where a single object is repeated across the grid in different patterns. Similar to the reuse task~\ref{subsubsubsec:reuse}, the objects used in the goal boards are modified versions of the skills generated by the Optimizer. In the repeat task, these objects are placed multiple times across the board in structured patterns, requiring the models to recognize and generalize the repetition. We are interested in: (a) whether the models can identify the correct repetition pattern, including spatial arrangement, order, and frequency, to accurately replicate it; and (b) whether the models can express this pattern using appropriate programming abstractions, such as loops or nested loops.

The repeat task follows the same two-player, multi-turn interaction as the reuse task. Cobot is equipped with the learned skill set and is expected to select the relevant skill, generate the corresponding function calls, and use suitable programming abstractions to place the object repeatedly across the grid. If Cobot cannot identify a suitable skill, Programmer provides step-by-step instructions, and the interaction falls back to the Teach reconstruction loop. The task is completed when Programmer responds with \texttt{DONE} or when the maximum turn limit ($30$) is reached. Once Programmer responds with \texttt{DONE}, GM evaluates task success by comparing the reconstructed board with the goal board at each cell location, including component color and stacking order.

\section{Experimental Setup}
\label{sec:expsetup}
As the TAA framework follows two-player interaction, both players are prompted in a zero-shot setting and receive the full interaction history at every turn. More details about the prompt context are available in the Supplementary Material.

\subsection{Dataset}
\label{subsec:dataset}
To evaluate the proposed approach, we require an environment that mimics a manipulation setting and allows collaboration over multiple turns. The tasks also require semantic representations expressed in code so that outputs can be evaluated objectively. We use the SARTCo~\citep{kranti2024towards} dataset for our experiments. The SARTCo dataset is an 8$\times$8 2.5D (a two-dimensional grid with vertical stacking) grid-world environment in which objects are arrangements of connected components of nuts, washers, screws, and bridges. A grid containing one or more such objects is referred to as a board. For the reconstruction and reuse tasks, the goal boards contain single objects, whereas the repeat task uses goal boards with multiple objects. Since the publicly released dataset contains a limited number of test samples ($130$), we generated a new set of samples using the same framework, producing a training set of 165 samples and validation and test sets of 495 samples each. The training set is used in the reconstruction task for teaching the skills. The validation and test sets contain three instances (that differ in colors and locations) for each of the $165$ unique skills in the training set without any overlap. This setup is designed to evaluate whether the learned skills can generalize to unseen variations.

\subsection{Ablation Studies}
\label{subsec:ablationstudies}
We conducted multiple ablation studies to design the prompt. This required several design decisions, including the number of turns before stopping the interaction, whether the cobot should ask clarification questions, whether execution error feedback improves reconstruction and overall success, and the number of turns allowed for the optimization step. Together, these decisions determine our final configuration. The detailed ablation results are available in the supplementary material.

\subsection{Evaluation with a Human Programmer}
\label{subsec:exphuman}
To evaluate the performance of the proposed pipeline under natural human instructions, we conducted a preliminary human-in-the-loop evaluation on a subset of the test set. We randomly selected 50 test instances (\texttt{seed = 76}) and employed a human participant as the Programmer, while the Cobot was evaluated on the reuse and repeat tasks. In these experiments, the Cobot was powered by \texttt{Qwen3-Coder} using the skill-set generated by \texttt{Qwen3-Coder}. The participant was a student assistant in the department who was familiar with the experimental setup but had not worked on this paper.

\subsection{Model Selection}
\label{subsec:modelselection}
We experiment with both an open-weight model (\texttt{Qwen3 Coder 30B A3B Instruct}~\footnote{\url{https://qwen.ai/blog?id=qwen3-coder}}) and a closed model (\texttt{GPT5.2-Codex}~\footnote{\url{https://developers.openai.com/api/docs/models/gpt-5.2-codex}}). These models were chosen based on their competitive performance on code generation benchmarks, representing both open-weight and closed model paradigms. We also investigate whether our approach can be realised with an open-weight model that can potentially be self-hosted and has latency advantages, by comparing it  to a proprietary model that incurs usage-based costs. Hence as a case-study, we evaluate one model from each paradigm. The experiments include using the same model for both the Programmer and the Cobot, as well as cross-model pairings (using \texttt{GPT5.2-Codex} as the Programmer and \texttt{Qwen3-Coder} as the Cobot, and vice versa). The rationale for the cross-model pairings is to understand the sensitivity of the Cobot to the choice of the Programmer and its impact on task success. The open-weight model is hosted locally on a single A100 GPU and is accessed through a local API endpoint. The closed model is accessed via Azure APIs. Both models are run with a temperature of $0$ to reduce output variance, with a \texttt{max\_new\_tokens} limit of $1000$ for reconstruction and reuse tasks and $2000$ for the repeat task.

\subsection{Evaluation Metrics}
\label{subsec:evalmetrics}
As the dataset provides a known target configuration along with its semantic representation, we conduct a detailed evaluation of the model responses. 

\textbf{Reconstruction task:} We use execution success to compare the reconstructed board state with the goal board in terms of the occupied cells (element type, color, location, and stacking order). In addition, we compute dialogue-level metrics, including success rate and clarification and correction rates normalized by the number of elements in the object, and the number of turns required to complete the reconstruction.

\textbf{Optimization task:} We use execution success to compare the board state produced by the generated optimized function with the goal board in terms of the occupied cells, measuring optimization accuracy. We also compute several structural metrics using Python's abstract syntax tree (AST), including whether the generated code contains a valid function definition, the reduction in \texttt{put} calls relative to the input, and the presence of loops indicating abstraction rather than repetition. We further measure the bytecode length before and after optimization using Python's \texttt{dis} module to assess structural compression.

\textbf{Reuse and Repeat tasks:} For the reuse and repeat tasks, we use execution success to compare the generated board with the ground truth board in terms of the occupied cells (element type, color, location, and order), measuring task accuracy when using learned skills. In addition, we measure the rate at which the model selects the correct function from the available skill set and the rate at which it applies this function with the correct arguments. We also compute the token-based cost for both \textit{No-skills} and \textit{Skills-Used} settings across self-play and cross-model pairings, analysing the trade-off between task success and computational cost.

\section{Results}
\label{sec:results}
We present our results based on the research questions presented in Section \ref{sec:introduction}, covering the talk–teach interaction, skill generation through optimization, and the reuse, repeat of learned skills in new contexts, providing both quantitative and qualitative analysis.

\begin{table*}[h]
\centering
\setlength{\tabcolsep}{10pt}
\caption{Performance comparison between \texttt{Qwen3-Coder} and \texttt{GPT5.2-Codex} on reconstruction and optimization tasks, reporting success rates for each component and overall performance along with token cost; Overall = Reconstruction x Optimization success; Token cost is the total across all test instances.}
\label{tab:reconststats}
\vspace{6pt}
\begin{tabular}{ccccc}
\hline
\multirow{2}{*}{Model}           & \multicolumn{3}{c}{Success Rate}                                   & \multirow{2}{*}{Token Cost (\$)} \\ 
                                 & Reconstruction       & Optimization         & Overall              &                             \\ \hline \\
\texttt{Qwen3-Coder} & 0.69 & 0.93 &  0.64 & 1.41 \\ \\
\texttt{GPT5.2-Codex} & \textbf{0.93} & \textbf{1.00} & 0.93 & \textbf{10.43}        \\ \hline
\end{tabular}%
\end{table*}

\subsection{Reconstruction Task}
\label{subsec:resultsreconstruction}
\textit{Can we through this methodology teach new behaviors through multi-turn natural language dialogue?} This question examines the teach stage of the TAA pipeline, where the Cobot learns to reconstruct a goal board through multi-turn dialogue with the Programmer.

Table~\ref{tab:reconststats} shows the success rate of teaching new skills. We first compute the reconstruction success rate, where the Programmer collaborates with the Cobot to accurately reconstruct the board while generating the primitive code snippets. The results show that reconstruction success is higher for the closed model (\texttt{GPT5.2-Codex}) than for the open-weight model (\texttt{Qwen3-Coder}). Some of the failures in \texttt{Qwen3-Coder} are due to the model repeatedly responding with ``Yes, I know'' without generating code snippets, or continuing to ask clarification questions without correcting misplacements caused by using the wrong colors or incorrect locations. Although reconstruction success varies between the models, both models show higher optimization success rates for successful reconstructions. The token cost is computed based on the number of tokens used and pricing information from OpenRouter\footnote{\url{https://openrouter.ai/}}, and is reported for reference only\footnote{\texttt{Qwen3-Coder} is hosted on a local GPU; however, the token price is computed to show comparison with commercial model usage.} as pricing is subject to change by API providers.

\begin{figure*}[h]
\centering
\caption{Performance of \texttt{Qwen3-Coder} and \texttt{GPT5.2-Codex} as task complexity increases (measured by the number of elements per object). The plots show average number of clarification and correction turns, and overall success rate for reconstruction task with varying structural complexity.}
\label{fig:reconstdlgstats}
\vspace{6pt}
\includegraphics[width=\textwidth]{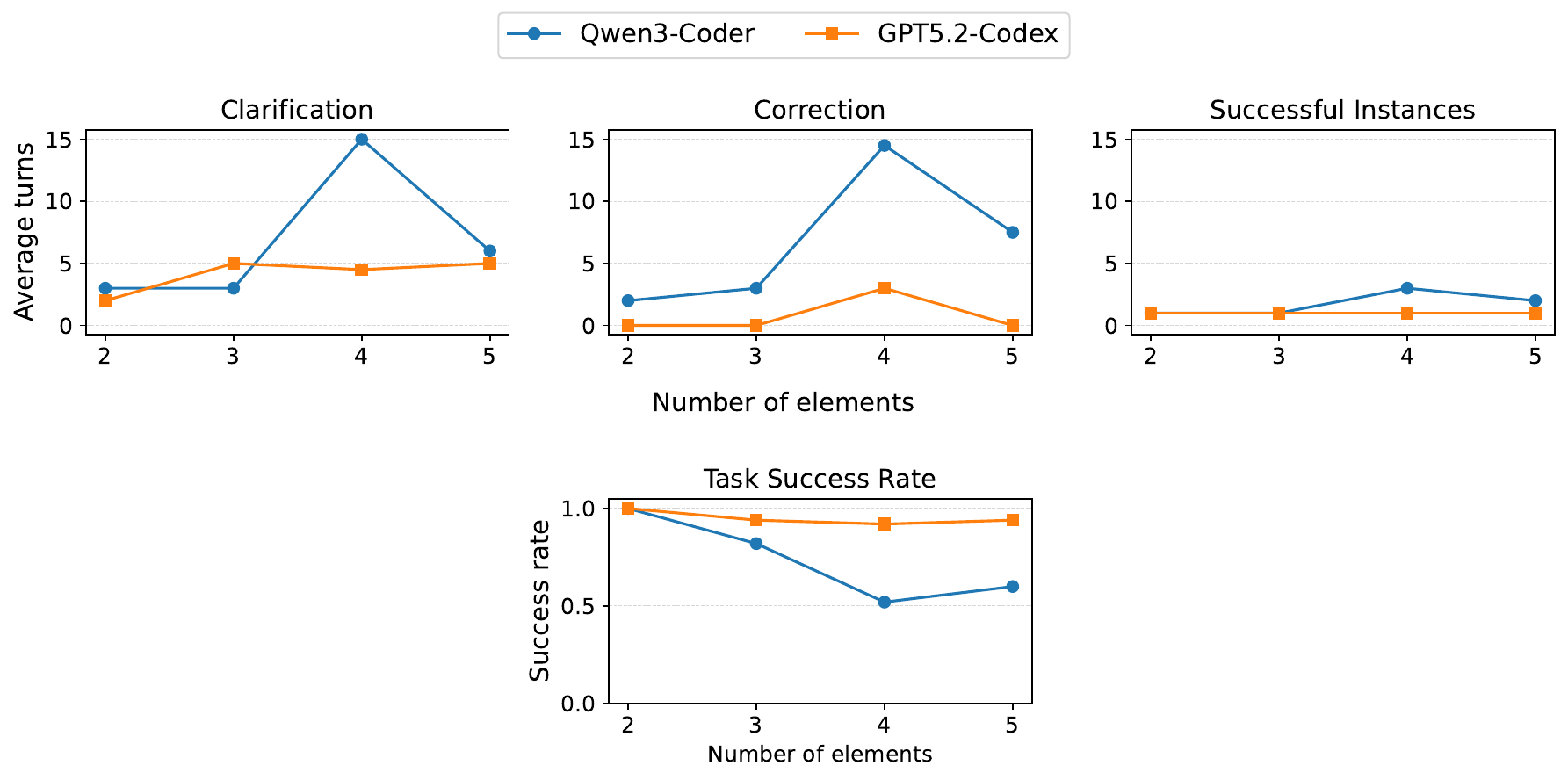}
\end{figure*}

These results indicate that it is possible to acquire new executable behaviors through the proposed framework, and that the success rate depends on the model's capabilities. We now 
examine the performance drops through the lens of object complexity and how it affects interaction behavior and reconstruction success. Specifically, we analyze how the success rate changes with the number of elements in the object, and whether clarification and correction rates change with increasing object complexity. Figure~\ref{fig:reconstdlgstats} shows that for \texttt{Qwen3-Coder}, clarifications and corrections appear to increase with the number of elements. In contrast, \texttt{GPT5.2-Codex} shows an almost flat trend in clarifications and only a few instances of corrections for objects with four elements, while all other cases show zero corrections. For both models, the reconstruction success rate declines as the number of elements in the object increases. The drop is higher for \texttt{Qwen3-Coder}, whereas \texttt{GPT5.2-Codex} appears less affected, although a slight decrease is still observed.

\subsubsection{Abstract: Reusable Function Generation}
After a successful Teach step, we now proceed with the Abstract step, to abstract the primitive action sequences. We evaluate \textit{how accurately models can abstract the given code snippets into a higher-order function}. 
\begin{figure*}[h]
\centering
\caption{Success rate of \texttt{Qwen3-Coder} and \texttt{GPT5.2-Codex} on code optimization objectives. Each point represents the proportion of test instances where the model correctly applied the required transformation, including using correct function names, using loop constructs, reducing \texttt{put()} calls, and minimizing output byte length; Vertical lines connect the two models scores to highlight the gap.}
\label{fig:optimstats}
\vspace{6pt}
\includegraphics[width=0.7\linewidth]{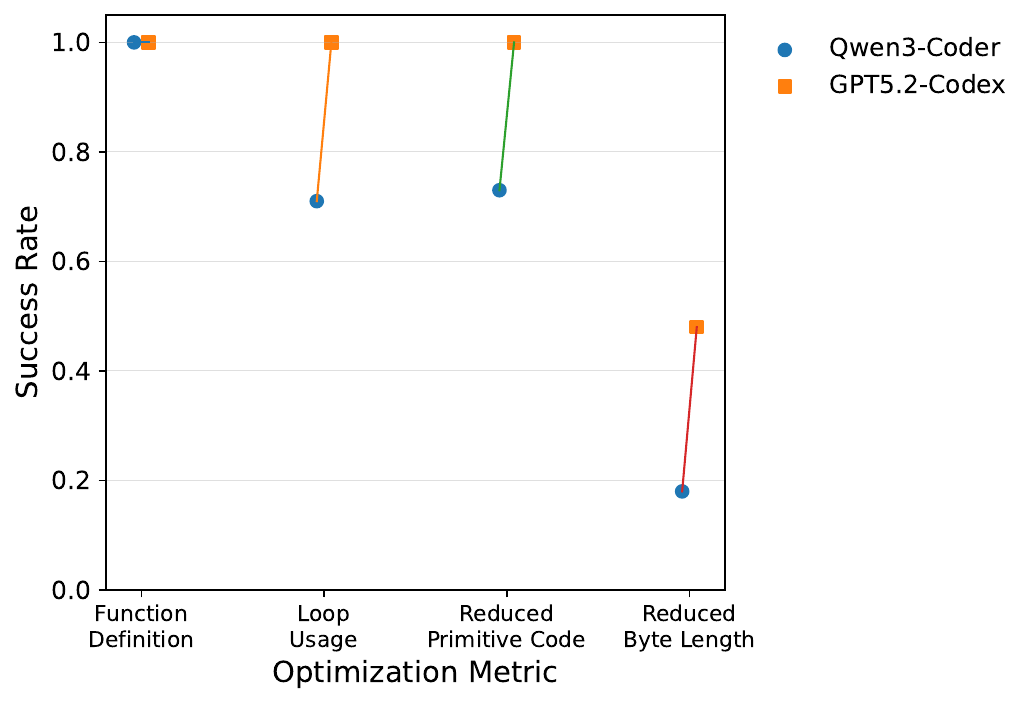}
\end{figure*}
Table~\ref{tab:reconststats} shows that both the models achieve higher optimization rates. That indicates the abstracted functions reproduce the behavior of the input primitive sequences. We next analyse how well these functions are structurally optimized. This question is important as the goal is not only successful reconstruction but also generating a function that is both optimized and reusable, capable of reproducing the learned behavior in new contexts. Figure~\ref{fig:optimstats} shows the analysis across model configurations.

We first check whether the output contains a valid function definition with the correct object name. We then examine how the primitive sequences are structured within the generated function. As the input consists of multiple \texttt{put()} calls, recurring patterns may be present, and an effective optimizer should abstract such patterns. We therefore evaluate whether a loop is present in the generated output and whether the overall number of \texttt{put()} calls is reduced, since loop alone does not guarantee correct abstraction. Finally, we compute whether the bytecode length is reduced after optimization. Both models consistently generate the correct function signature. \texttt{GPT5.2-Codex} effectively uses loops and reduces the number of \texttt{put()} calls, while \texttt{Qwen3-Coder} shows competitive performance on both metrics. However, neither model shows a significant reduction in bytecode length, although \texttt{GPT5.2-Codex} performs slightly better. This is because Python's loop constructs and data structures such as list definitions, zip, and tuple unpacking introduce compilation overhead that can offset or exceed the savings from fewer \texttt{put()} calls. Overall, this shows that the optimizer able to abstract the code that is not only functionally correct but also structurally optimized.

Taken together, the results suggest that Teach and Abstract steps enable the acquisition of new executable behaviors through multi-turn natural-language dialogue and their abstraction into reusable skills. The generalisability of these abstracted skills is further evaluated through the reuse and repeat tasks.

\subsection{Reuse}
\label{subsec:results-reuse}
With the generation of reusable skills, the next step is to evaluate whether these skills can be applied in new scenarios. We evaluate three settings: the test set without any skills (\textit{no-skills}), which effectively reduces the reuse task to a reconstruction-style test and acts as the baseline; the test set using the skill set learned by \texttt{Qwen3-Coder} (\textit{Qwen skills}); and finally using the skill set learned by \texttt{GPT5.2-Codex} (\textit{Codex skills}). This allows us to study the impact of having learned skills on overall task success and associated cost. We additionally evaluate cross-pairings of the models to examine whether the choice of Programmer (instruction giver) influences the Cobot (instruction follower) model. In cases where a corresponding skill is unavailable, as not all reconstruction instances are successful, these instances fall back to the reconstruction-style task.

\begin{figure*}[h]
\centering
\caption{Cost–performance trade-off for the reuse task. Each point represents a model configuration, comparing success rate and token cost when responses are generated without skills, with Qwen skills, or with Codex skills; G-Q, Q-Q etc. Labels denote Programmer–Cobot pairs; G-Q: GPT as Programmer and Qwen as Cobot; Q-G: Qwen as Programmer and GPT as Cobot; The y-axis is truncated to the range 0.6-1.0 to highlight differences among performance scores.}
\label{fig:reusestats}
\vspace{6pt}
\includegraphics[width=\textwidth]{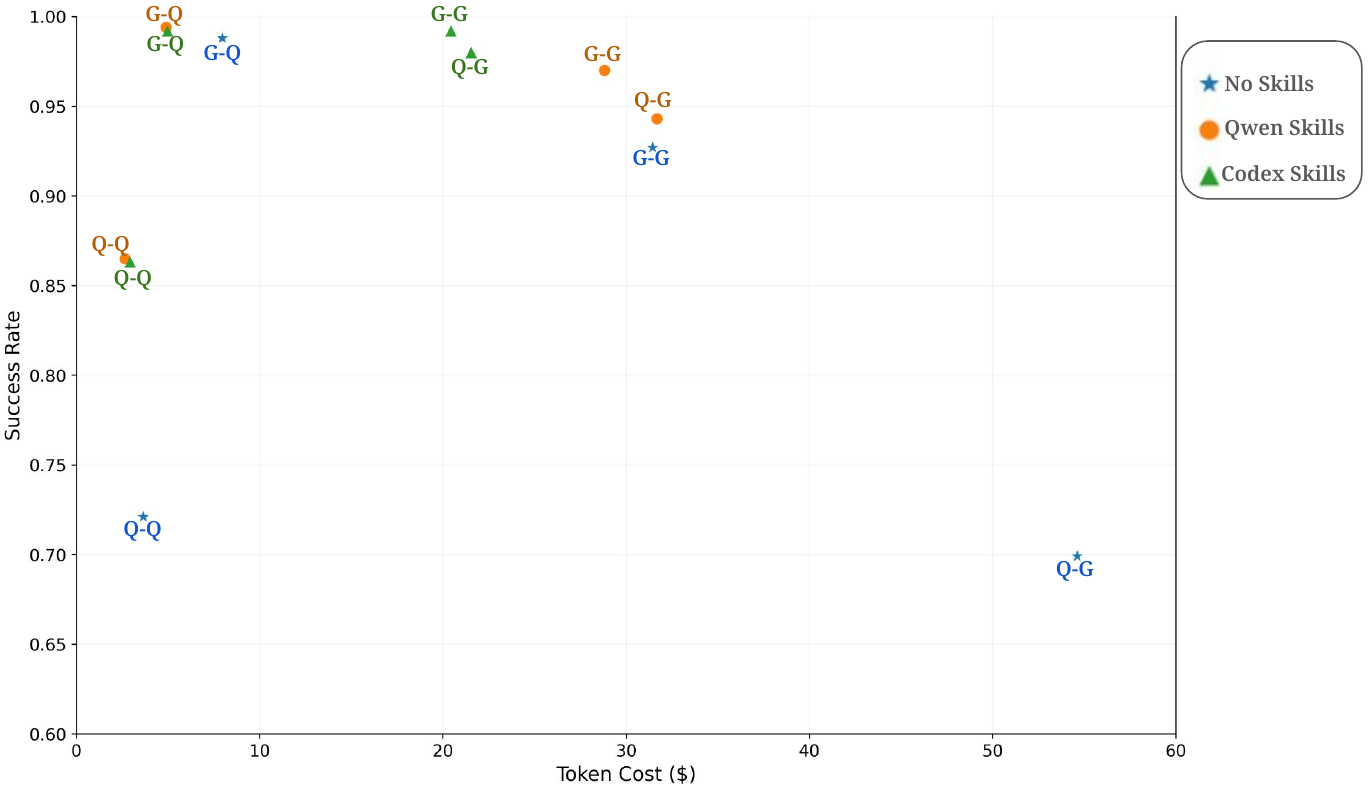}
\end{figure*}

\textbf{Impact of Skill Usage:} As shown in Figure~\ref{fig:reusestats}, \textit{using learned skill sets improves task success compared to the no-skill baseline}. For the \texttt{Qwen3-Coder} Cobot using the same model as programmer (Q–Q), the success rate increases from 0.72 without skills to 0.86 when using the \texttt{Qwen3-Coder} skill set and to 0.87 when using the \texttt{GPT5.2-Codex} skill set. When the programmer is changed to \texttt{GPT5.2-Codex} (G–Q), the success rate increases to 0.99 for both skill sets (Qwen3 skills: 0.99; Codex skills: 0.99). We further observe that using skill sets reduces overall token cost compared to the no-skill baseline. Without skills, the cost is \$3.63 for Q–Q and \$7.96 for G–Q. With skill reuse, the cost becomes \$2.64 for Q–Q and \$4.88 for G–Q when using Qwen3 skills, and \$2.92 for Q–Q and \$4.96 for G–Q when using Codex skills.

While the \texttt{Qwen3-Coder} Cobot shows a larger improvement in success rate with the use of skills (Q–Q), the \texttt{GPT5.2-Codex} Cobot (G–G) shows a smaller increase, though with a more substantial cost reduction. For \texttt{GPT5.2-Codex}, the success rate increases from 0.93 without skills to 0.97 when using the Qwen3 skill set and to 0.99 when using the Codex skill set. The reduction in cost is larger, decreasing from \$31.44 without skills to \$28.82 with the Qwen3 skill set and to \$20.44 with the Codex skill set. A similar trend is observed for the cross-pairing setting (Q–G), where the \texttt{GPT5.2-Codex} Cobot benefits less from cross-pairing, with the success rate decreasing and cost remaining higher. In particular, the cost reaches \$54.62 in the no-skill setting compared to \$31.44 for G–G. This behavior is largely due to ambiguous instructions that trigger additional clarification questions, resulting in longer interactions, increased token usage, and slightly lower reconstruction success.

These results demonstrate that \textit{skills generated} during the reconstruction task are \textit{model-agnostic} and can be reused by cobots powered by different models for subsequent tasks resulting in \textit{improved success rates and reduced costs}. We next evaluate the applicability of these reusable skills in the Repeat test.

\subsection{Repeat}
\label{subsec:results-repeat}
Similar to the reuse task evaluation, the repeat task examines whether the models can use learned skills to reconstruct boards that contain repeated placements of the same object across the grid. The experiments consider three settings: (a) no-skills, where the models perform the task without access to any learned skills; (b) \texttt{Qwen3-Coder} skills, a skill set learned by the \texttt{Qwen3-Coder} Cobot; and (c) \texttt{GPT5.2-Codex} skills, a skill set learned by the \texttt{GPT5.2-Codex} Cobot. The models are evaluated in both self-play and cross-pairing settings.

\begin{figure*}[h]
\centering
\caption{Trade-off between token cost and repeat task success rate across model configurations. Results compare responses generated without skills, with Qwen skills, and with Codex skills, showing that skill usage significantly improves success rates while maintaining relatively low token cost; The y-axis is truncated to the range $0.4-1.0$ to highlight differences among performance scores.}
\label{fig:repeatstats}
\vspace{6pt}
\includegraphics[width=\textwidth]{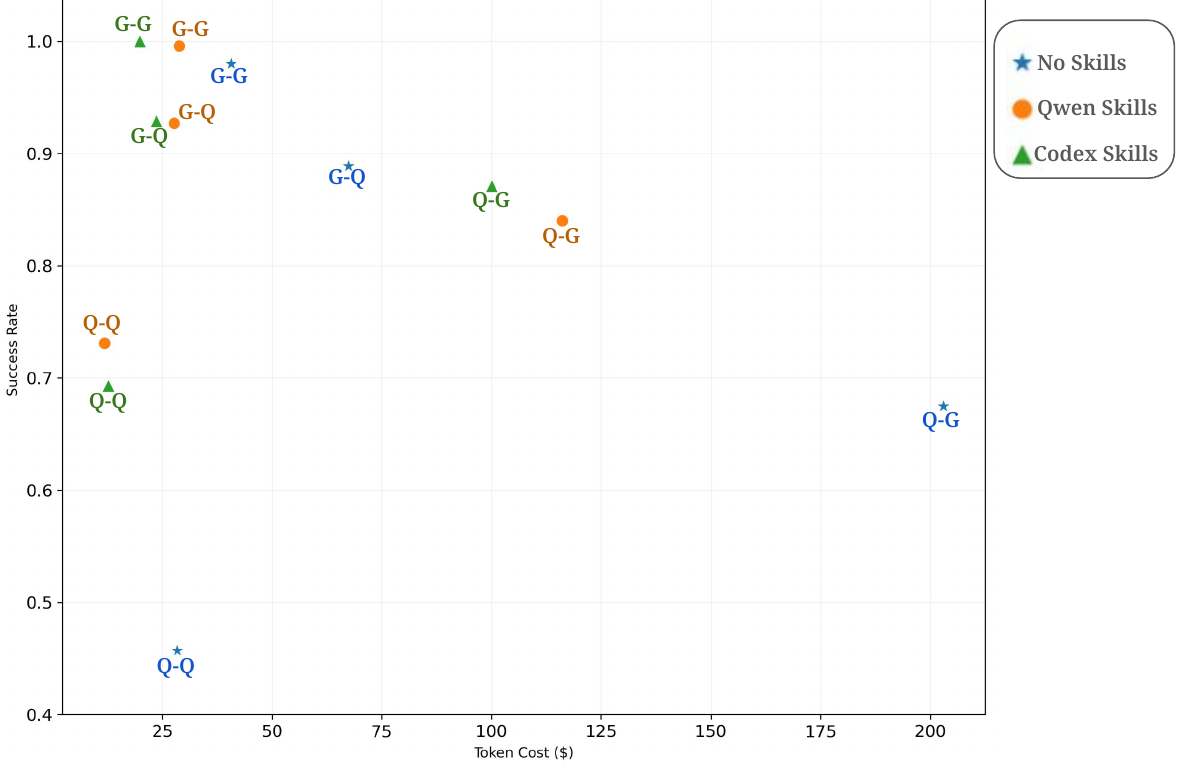}
\end{figure*}

\textbf{Impact of Skill Usage:} As shown in Figure~\ref{fig:repeatstats}, the repeat task also shows benefits from the use of learned skills for the \texttt{Qwen3-Coder} Cobot. In self-play, the success rate increases from 0.46 without skills to 0.73 and 0.70 when using the \texttt{Qwen3-Coder} and \texttt{GPT5.2-Codex} skill sets respectively, while the cost is reduced from \$28.37 to \$11.84 and \$12.70. When cross-paired with \texttt{GPT5.2-Codex} as the Programmer, the success rate increases from 0.89 to 0.93 for both skill sets, while the cost decreases from \$67.40 to \$27.68 and \$23.65. The higher cost in the repeat task compared to other tasks is due to repeated placements of the same object, which lead to multiple instructions, clarification requests, or corrections, and therefore increased token usage.

In contrast, the \texttt{GPT5.2-Codex} Cobot shows similar success rates across the no-skill and skill settings. The success rate changes from 0.98 without skills to 0.97 with the \texttt{Qwen3-Coder} skill set and 0.99 with the \texttt{GPT5.2-Codex} skill set, while the cost decreases from \$40.67 to \$28.89 and \$19.90 respectively. The decrease in success rate in the cross-pairing case is due to ambiguous instructions, similar to the reuse scenarios. This effect is more in the repeat task compared to the reuse task because the reuse task involves a single object placement, whereas the repeat task requires placing the same object multiple times on the grid.

Together, the Reuse and Repeat evaluations show that the skills learned from Teach and Abstract steps can be Applied to new contexts across different model configurations. The learned skills are transferable across models, can improve task performance, and reduce the associated costs.

\subsection{Skill Retrieval Analysis}
\label{subsec:results-skillretrievalanalysis}
Sections~\ref{subsec:results-reuse} and~\ref{subsec:results-repeat} show that skill usage improves the overall task performance, we next analyse how much skill use contributed to success and how much success comes from the fallback reconstruction flow.

\begin{table*}[h]
\centering
\setlength{\tabcolsep}{10pt}
\caption{Skill retrieval rate across model configurations for Reuse and Repeat tasks. The highlighted values indicate the higher-performing configuration when using the corresponding skill set. Labels denote Programmer–Cobot pairs; G-Q: GPT as Programmer and Qwen as Cobot; Q-G: Qwen as Programmer and GPT as Cobot; OTSR: Overall Task Success Rate; SRSR: Skill-retrieval Success Rate; FRSR: Fall-back Reconstruction Success Rate. } 
\label{tab:reuse-skillretrievalrate}
\vspace{6pt}
\begin{tabular}{cccccc}
\hline \\
Evaluation Task & Skill-Set & Model & OTSR & SRSR (\%) & FRSR (\%) \\ \\ \hline \\
\multirow{8}{*}{Reuse} & \multirow{4}{*}{Qwen} & Q-Q   & 0.865 & 83.64 & 16.35    \\
                       &                       & G-Q   & 0.994 & \textbf{93.90} & 6.10     \\
                       &                       & G-G   & 0.970 & 86.67 & 13.33    \\
                       &                       & Q-G   & 0.943 & 88.65 & 11.35    \\\\
                      & \multirow{4}{*}{Codex} & Q-Q & 0.863 & 89.23 & 10.77    \\
                      &                        & G-Q   & 0.992 & 94.50 & 5.50    \\
                      &                        & G-G   & 0.992 & \textbf{97.35} & 2.65 \\
                      &                        & Q-G   & 0.980 & 96.70 & 3.30    \\\\ \hline \\

\multirow{8}{*}{Repeat} & \multirow{4}{*}{Qwen} & Q-Q   & 0.731 & \textbf{96.13} & 3.87    \\
                        &                     & G-Q   & 0.927 & 93.25 & 6.75     \\
                        &                     & G-G   & 0.996 & 81.95 & 18.05    \\
                        &                     & Q-G   & 0.840 & 88.22 & 11.78    \\\\
                       & \multirow{4}{*}{Codex} & Q-Q & 0.693 & \textbf{98.25} & 1.75    \\
                       &                      & G-Q   & 0.929 & 96.30 & 3.70    \\
                       &                      & G-G   & 1.000 & 96.97 & 3.03 \\
                       &                      & Q-G   & 0.871 & 96.52 & 3.48    \\ \hline

\end{tabular}%
\end{table*}

\subsubsection{Skill Retrieval Rate}
\label{subsubsec:skillretrievalrate} 
\textbf{Reuse Task:} Table~\ref{tab:reuse-skillretrievalrate} show the results. These results indicate that models retrieve the correct skill from context in most cases, and that the \texttt{Qwen3-Coder} Cobot is more sensitive to the choice of Programmer and has more impact on the retrieval rate than on \texttt{GPT5.2-Codex}, although a slight decrease is also observed for \texttt{GPT5.2-Codex} in cross-pairing settings. Moreover, irrespective of the number of available skills (Qwen skill set: 315; Codex skill set: 415), the task success rate and skill retrieval rate are similar across skill sets for the \texttt{Qwen3-Coder} Cobot: 0.865, 83.64\% for the Qwen skill set, and 0.863, 89.23\% for the Codex skill set. The drop in success rates are likely due to difficulty in extracting the correct skill name from instructions and matching it with the available skill set.

\textbf{Repeat Task:} Similar to the reuse task, we further analyse how skill usage affects the repeat success rate. Table~\ref{tab:reuse-skillretrievalrate} shows these statistics across model configurations using the Qwen skill set and the Codex skill set. Unlike Reuse task, the skill retrieval rate is similar for the \texttt{Qwen3-Coder} Cobot, irrespective of the Programmer and the skill-set. However, its overall task success rate improves to 0.927, 0.929 for Qwen and Codex skill set respectively, when paired with the \texttt{GPT5.2-Codex} Programmer. In contrast, the overall task performance of the \texttt{GPT5.2-Codex} Cobot drops from 0.996 to 0.840 for the Qwen skill set and from 1.000 to 0.871 for the Codex skill set in cross-pairing. This once again highlights the sensitivity of the Cobot to the Programmer used and its impact on overall task performance.

Overall, these results suggest that the models retrieve the correct skill from context in most cases and the skill retrieval rates drive the overall task success rates across model configurations.

\begin{table*}[h]
\centering
\setlength{\tabcolsep}{10pt}
\caption{Skill hallucination percentages across model configurations for Reuse and Repeat tasks computed as the ratio of hallucinated skills to the total retrieved skills. Labels denote Programmer–Cobot pairs; G-Q: GPT as Programmer and Qwen as Cobot; Q-G: Qwen as Programmer and GPT as Cobot; } 
\label{tab:skillhallucination}
\vspace{6pt}
\begin{tabular}{cccc}
\hline \\
Evaluation Task & Skill-Set & Model & Hallucinated (\%) \\\\ \hline \\
\multirow{8}{*}{Reuse} & \multirow{4}{*}{Qwen} & Q-Q   & 29.61    \\
                       &                       & G-Q   & 34.41    \\
                       &                       & G-G   & 24.76    \\
                       &                       & Q-G   & 26.09    \\\\
                      & \multirow{4}{*}{Codex} & Q-Q   & 7.87   \\
                      &                        & G-Q   & 6.47    \\
                      &                        & G-G   & 4.18 \\
                      &                        & Q-G   & 3.62    \\\\ \hline \\

\multirow{8}{*}{Repeat} & \multirow{4}{*}{Qwen} & Q-Q & 31.89    \\
                        &                     & G-Q   & 33.18     \\
                        &                     & G-G   & 22.28   \\
                        &                     & Q-G   & 25.34   \\\\
                       & \multirow{4}{*}{Codex} & Q-Q & 7.12    \\
                       &                      & G-Q   & 5.87    \\
                       &                      & G-G   & 3.75 \\
                       &                      & Q-G   & 3.85    \\ \hline

\end{tabular}%
\end{table*}

\subsubsection{Skill Hallucination:} Skill retrieval analysis in Section~\ref{subsubsec:skillretrievalrate} shows that the models are able to retrieve most of the skills across different configurations. However, during a random manual analysis of skill usage, we observed scenarios, where the model generates a skill that is not available (the skill was not learned during reconstruction) in the model context. We refer to these scenarios as \textit{skill hallucinations} and analyse their impact on the task success. However, these are not random hallucinations as the generated function call follows the correct function signature, with the correct skill name and relevant arguments. As a design choice, we execute these generated skills in the virtual simulator using the ground truth function definition. This choice allows us to isolate and evaluate the Cobot's ability to correctly predict skill usage, specifically the correct object name, colors, locations, regardless of whether the skill exists in the library.

Table~\ref{tab:skillhallucination} shows the percentage of hallucinated skills across skill sets and model configurations. The hallucination percentages are higher when the Qwen skill set is used, as only 315 skills are available for the 495 test instances, compared to the Codex skill set, for which 415 skills are available. We also compute the success rate for these hallucinated instances and the success rate is $1.0$ regardless of the model configuration, further confirming that the models generally predict task-relevant skills.

Overall, these results demonstrate that \textit{skills generated} during the reconstruction task are \textit{model-agnostic} and consistently improve task performance when successfully retrieved. The impact of skill reuse varies by model: for \texttt{Qwen3-Coder}, \textit{skill reuse primarily improves success rates}, while for \texttt{GPT5.2-Codex}, \textit{the benefit is more in terms of cost reduction}. Skill retrieval rates are high across all configurations. In cases where skills are unavailable, models tend to hallucinate skills that align with the context. Regardless of hallucination, the predicted skills align with the input context, confirming that the models generally predict task-relevant skills. When retrieval doesn't succeed, the interaction falls back to reconstruction, where success rates vary considerably across model configurations. We also observe that both Cobots are sensitive to the choice of Programmer and this has an impact on retrieval rates. Together, these findings highlight that \textit{reliable skill retrieval and context-aware skill prediction become increasingly important as task complexity grows}.

\subsection{Qualitative Analysis}
\begin{figure*}[h]
\centering
\caption{Qualitative examples (Q-Q setup) showing clarification and correction during reconstruction, misinterpretation during skill reuse, fallback from reuse to reconstruction, and hallucinated skill use. These showcase how Programmer and Cobot interact across task settings, how errors arise from ambiguous or incorrect skill selection, and how the system either continues through correction, falls back to primitive reconstruction; Q-Q: \texttt{Qwen3-Coder} is used as both Programmer and Cobot.}
\label{fig:qualitativeanalysis}
\vspace{6pt}
\includegraphics[width=\textwidth]{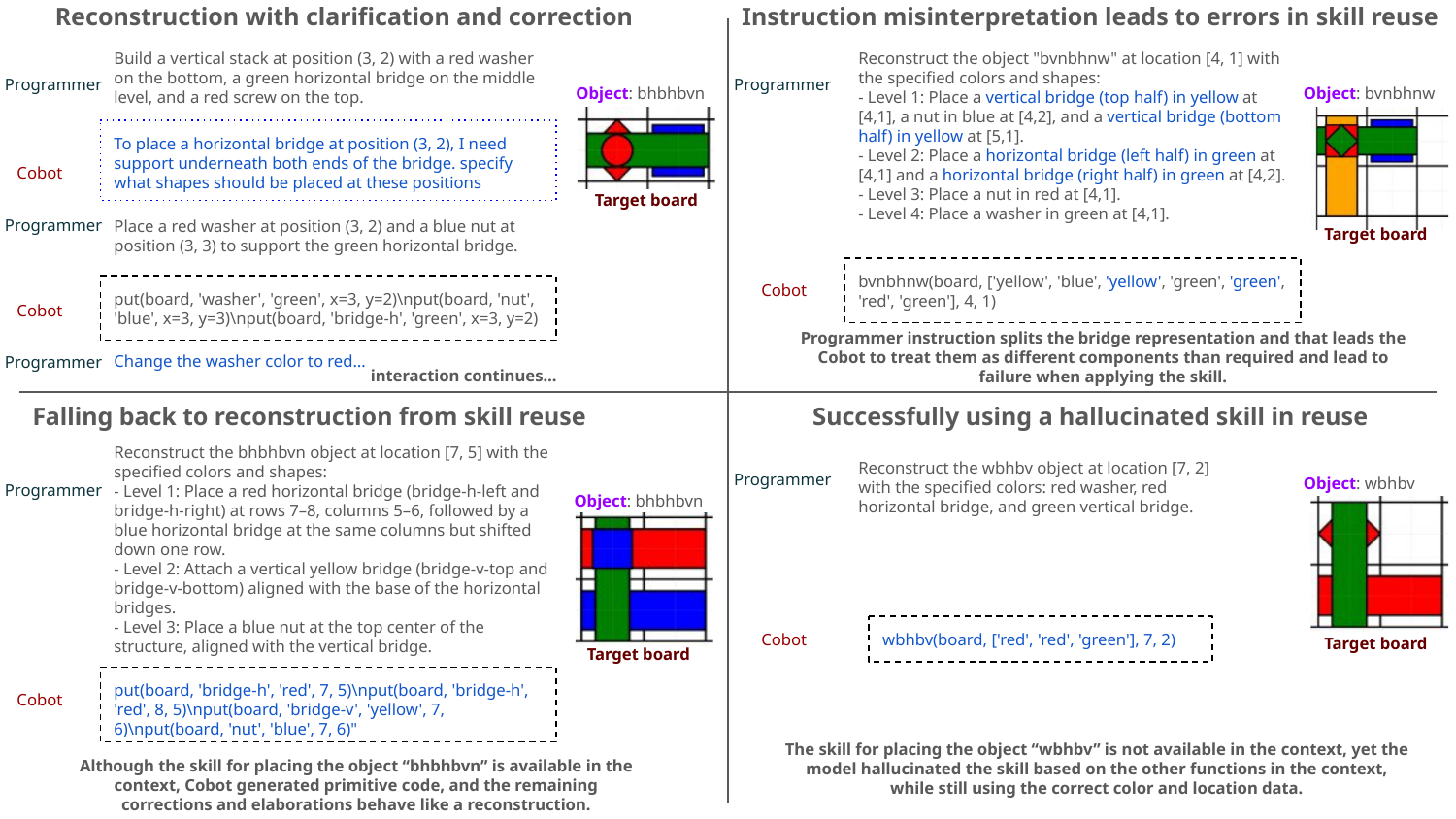}
\end{figure*}

The qualitative examples in Figure~\ref{fig:qualitativeanalysis} highlight different behaviors across the reconstruction and reuse settings. In reconstruction, the interaction supports clarification and correction. When Cobot cannot infer the instruction, it asks Programmer for additional information. Programmer then elaborates the instruction, and later corrects an error in the generated code by specifying that the washer should be red. This example shows how the multi-turn setup allows the task to continue even when the initial instruction is incomplete or when the generated code contains a mismatch.

The reuse examples show that skill availability alone does not guarantee correct skill use. In one case, Cobot misinterprets the object description by treating the two halves of a bridge as separate components, which leads to an incorrect function call. This suggests that reuse errors can also arise from Cobot interpretations. In another case, although the relevant skill is available in the context, Cobot fails to call it and instead generates primitive code. The interaction then falls back to the reconstruction loop, where Programmer continues with step-by-step instructions and corrections.

The reuse scenario also showcases a different failure mode, where Cobot calls a hallucinated skill that is not present in the available skill library. However, the generated call still uses the correct colors and location arguments, indicating that the model can sometimes infer the intended structure even when it is not available in the context. Together, these cases suggest that TAA’s multi-turn design helps recover from ambiguity and local execution errors, while reuse failures often arise from skill grounding issues, instruction misinterpretation, and incorrect selection or application of learned skills.

\begin{figure*}[h]
\centering
\caption{Reuse Task: Clarification behavior under skill reuse settings as a function of task complexity.
(a) Clarification episodes remain sparse across all model configurations, with occasional increases for G–Q at higher complexity, while Q–Q and H–Q exhibit minimal clarification needs overall.
(b) Median dialogue turns for successful instances are nearly constant, with H–Q consistently requiring two turns, and both Q–Q and G–Q completing interactions in approximately a single turn.
Model Configuration: Q–Q (\texttt{Qwen3-Coder}–\texttt{Qwen3-Coder}), G–Q (\texttt{GPT5.2-Codex}–\texttt{Qwen3-Coder}), and H–Q (Human–\texttt{Qwen3-Coder}).}
\label{fig:reusehumanstats}
\vspace{6pt}
\includegraphics[width=\textwidth]{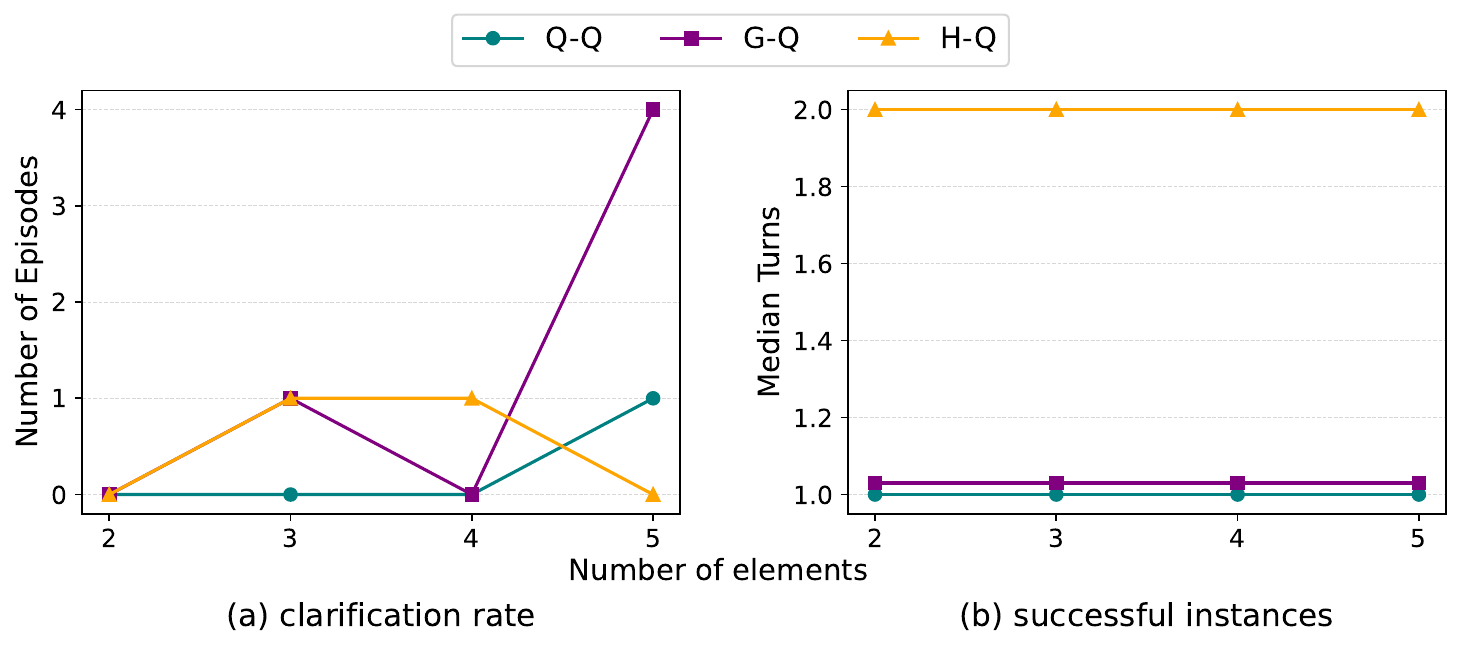}
\end{figure*}

\begin{figure*}[h]
\centering
\caption{Repeat Task: Clarification behavior across model configuration as a function of task complexity.
(a) Number of clarification episodes increases with the number of elements, with Q–Q and H–Q showing stronger sensitivity compared to G–Q.
(b) Median dialogue turns for successful instances remain relatively stable, with H–Q requiring consistently more turns, while G–Q maintains minimal interaction length across all settings.
Model Configuration: Q–Q (\texttt{Qwen3-Coder}–\texttt{Qwen3-Coder}), G–Q (\texttt{GPT5.2-Codex}–\texttt{Qwen3-Coder}), and H–Q (Human–\texttt{Qwen3-Coder}).}
\label{fig:repeathumanstats}
\vspace{6pt}
\includegraphics[width=\textwidth]{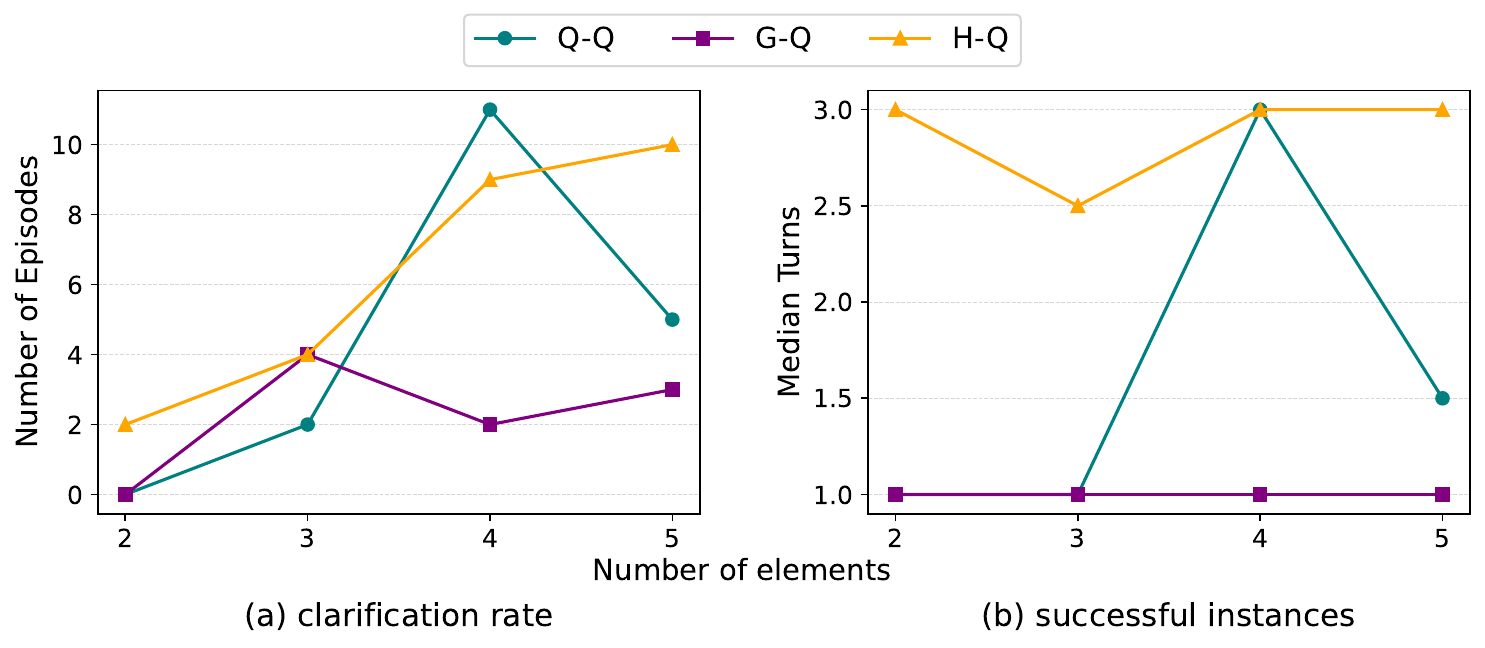}
\end{figure*}

\begin{table*}[h]
\centering
\setlength{\tabcolsep}{10pt}
\caption{Success rates on reuse and repeat tasks across model configurations, including a human baseline for the evaluated subset using Qwen skill set. Results highlight how configuration choices impact performance, with best results per task shown in bold.}
\label{tab:humanstudy}
\vspace{6pt}
\begin{tabular}{ccc}
\hline
Model Configuration & Reuse & Repeat \\ \hline
Human-Q            &  0.96     &  \textbf{0.94}     \\ \\
Q-Q               &  0.88     &  0.78  \\ \\
G-Q               &  \textbf{1.00} & \textbf{0.94}   \\ \hline
\end{tabular}%
\end{table*}

\subsection{Evaluation with a Human Programmer}
\label{subsec:humanstudies}
As mentioned in Section~\ref{subsec:exphuman}, we randomly selected $50$ test instances for the reuse and repeat tasks and evaluated with one human Programmer as a pilot evaluation. This evaluation (see Table~\ref{tab:humanstudy}) measures the performance of the Apply step when natural language instructions are provided by a human Programmer and the Cobot is powered by \texttt{Qwen3-Coder}. Overall, the Cobot is able to interpret human instructions and achieves comparable success rates to the model-based Programmer configurations for both the reuse and repeat tasks. The human Programmer achieves higher success rates than the \texttt{Qwen3-Coder} Programmer in the same Cobot configuration, suggesting that the Cobot is able to interpret human instructions effectively.

We analyse the episodes where clarification questions are present and whether the interaction behavior differs between human and model-based Programmers. Figure~\ref{fig:reusehumanstats} plot (a) shows how many episodes contain clarification questions for each model configuration in the reuse task. Q-Q has only one instance with a clarification question for a 5-element object, while G-Q has four instances for 5-element objects. H-Q has one episode each for three and four element objects. For the repeat task, the number of clarification episodes is slightly higher, with Q-Q peaking at 10 episodes for 4-element objects. We also analyse the median number of turns for all successful instances for each configuration. For the reuse task, the human Programmer consistently takes two turns while model-based Programmers take one turn. For the repeat task, the human Programmer takes three turns compared to one turn for model-based Programmers, suggesting that human instructions require more interaction to complete successfully.

Although this is a pilot study and the number of test instances are limited, this evaluation suggests that the pipeline can be extended to human Programmers, and that the Cobot can retrieve and apply skills even when instructions are human-written rather than generated by another model.

\section{Conclusion}
In this work, we propose the Teach-Abstract-Apply (TAA) framework, which enables users to teach collaborative robots new spatial manipulation routines through multi-turn natural language dialogue. The framework defines a three-stage pipeline: Teach, where a user interacts with the robot and iteratively refines the workflows; Abstract, where the resulting interaction trace is abstracted into a generalised behavior (skill); and Apply, where the abstracted skills are applied in different settings. We then test the TAA framework through a collaborative structure-building task, in which a Programmer instructs the Cobot to reconstruct a goal board through dialogue, the resulting interaction is abstracted into a reusable Python skill function and these abstracted skills are applied to new scenarios through reuse and repeat tasks. Our evaluation using both open-weight (\texttt{Qwen3-Coder}) and closed (\texttt{GPT5.2-Codex}) models in self-play and cross-model configurations on the SARTCo dataset demonstrates that:
\begin{itemize}
\item Abstracted skills are model-agnostic and consistently improve task success rates while reducing interaction cost compared to the no-skill baseline.
\item Skill retrieval rates are high across all configurations, with models demonstrating strong context-awareness in identifying and applying relevant skills.
\item Models tend to hallucinate skills that align with the task context, and these hallucinated skills achieve a 1.0 success rate, confirming that models correctly predict skill usage even when the exact skill is absent from the library.
\item The benefit of skill reuse varies by model capability, with smaller models benefiting more in terms of success rate improvement and larger models benefiting more in terms of cost reduction.
\item A preliminary human evaluation suggests that the pipeline also works with human instructions without requiring changes to the system.
\end{itemize}
These findings demonstrate the potential of combining dialogue-based teaching with program synthesis for scalable, reusable robot skill learning. Future work will focus on expanding the human evaluation to a more diverse participant pool, and evaluating the framework on diverse tasks and modalities. Additionally, incorporating safety mechanisms for real-world deployment and exploring the integration of TAA with a physical robot arm remain important directions for future work.

\section{Limitations}

Our proposed TAA framework demonstrates promising results for dialogue-based robot skill acquisition, but several limitations remain.

\textbf{Physical Robot:} Our proposed setup uses a virtual simulator in a controlled setting. As a design choice, we did not include the physical robot in the current setup, as perception, motion planning and low-level control may introduce additional failures~\citep{10.1145/3757279.3785550} beyond these associated with TAA components. We view this as a realistic starting point for evaluating the feasibility of TAA independently, with the potential to extend it to real-world settings involving a physical robot.

\textbf{Evaluation of Skill Quality:} The optimizer's correctness is currently judged by board state reproduction and structural metrics. While execution success confirms robustness and structural metrics assess minimality, the current evaluation does not assess higher-order properties such as skill composability beyond the reuse and repeat tasks.

\textbf{Baselines:} The primary comparison is against a no-skill baseline within the TAA framework. To our knowledge, no existing baseline combines dialogue-based teaching with persistent skill reuse, making direct comparison with prior work challenging.

\textbf{Safety:} The current framework operates in a controlled simulation environment. Scaling to real-world deployment would require additional safety mechanisms, such as ambiguity detection, formal verification of generated code, or constrained APIs with whitelisted operations, to mitigate risks associated with hallucinated skills or ambiguous instructions.

\textbf{Dataset Scale:} The experiments are conducted on the SARTCo dataset, which covers a specific grid-world environment. Evaluating TAA on larger and more diverse datasets with varied object types, spatial configurations, and task complexities would strengthen the generalizability of the findings.

\textbf{Model Dependence:} The results show that the benefit of skill reuse depends on the underlying model. Smaller open-weight models such as \texttt{Qwen3-Coder} benefit more in terms of success rate improvement, while larger closed models such as \texttt{GPT5.2-Codex} benefit more in terms of cost reduction. This suggests that the framework's effectiveness may vary with model capability.

\textbf{Human Evaluation Scale:} The human-in-the-loop evaluation was conducted as a pilot study with a single participant, limiting the statistical generalizability of the findings. A broader user study with a larger and more diverse participant pool, incorporating objective measures such as turns-to-success, clarification rates, and task completion time, remains an important direction for future work.

\section*{Conflict of Interest Statement}

The authors declare that the research was conducted in the absence of any commercial or financial relationships that could be construed as a potential conflict of interest.

\section*{Author Contributions}
CK: coded the framework, analysed the results and wrote the original draft.; SH: contributed to the conceptualization and reviewed and edited parts of the paper. ; DS: conceived the idea, managed the project and reviewed and edited parts of the paper.

\section*{Funding}
Kranti has been funded by the Bundesministerium für Bildung und Forschung (BMBF, German Federal Ministry of Research), project "COCOBOTS" (01IS21102A) and Deutsche Forschungsgemeinschaft (DFG, German Research Foundation) grant 423217434 (“RECOLAGE”), Completion Scholarship from Department of Linguistics and the Gertrud-Feiertag-Stipendium from the University of Potsdam.

\section*{Acknowledgments}
This is a short text to acknowledge the contributions of specific colleagues, institutions, or agencies that aided the efforts of the authors.

\section*{Supplemental Data}
 \href{http://home.frontiersin.org/about/author-guidelines#SupplementaryMaterial}{Supplementary Material} should be uploaded separately on submission, if there are Supplementary Figures, please include the caption in the same file as the figure. LaTeX Supplementary Material templates can be found in the Frontiers LaTeX folder.

\section*{Data Availability Statement}
The data, framework implementation and other resources are available at:\\ \url{https://github.com/clp-research/cobotprogsynthesis-coderep}.

\bibliographystyle{Frontiers-Harvard} 
\bibliography{test}

@inproceedings{kranti-etal-2025-clem,
    title = "clem:todd: A Framework for the Systematic Benchmarking of {LLM}-Based Task-Oriented Dialogue System Realisations",
    author = "Kranti, Chalamalasetti  and
      Hakimov, Sherzod  and
      Schlangen, David",
    editor = "B{\'e}chet, Fr{\'e}d{\'e}ric  and
      Lef{\`e}vre, Fabrice  and
      Asher, Nicholas  and
      Kim, Seokhwan  and
      Merlin, Teva",
    booktitle = "Proceedings of the 26th Annual Meeting of the Special Interest Group on Discourse and Dialogue",
    month = aug,
    year = "2025",
    address = "Avignon, France",
    publisher = "Association for Computational Linguistics",
    url = "https://aclanthology.org/2025.sigdial-1.5/",
    pages = "62--92"
}

@article{DBLP:journals/corr/abs-2310-03714,
  author       = {Omar Khattab and
                  Arnav Singhvi and
                  Paridhi Maheshwari and
                  Zhiyuan Zhang and
                  Keshav Santhanam and
                  Sri Vardhamanan and
                  Saiful Haq and
                  Ashutosh Sharma and
                  Thomas T. Joshi and
                  Hanna Moazam and
                  Heather Miller and
                  Matei Zaharia and
                  Christopher Potts},
  title        = {DSPy: Compiling Declarative Language Model Calls into Self-Improving
                  Pipelines},
  journal      = {CoRR},
  volume       = {abs/2310.03714},
  year         = {2023},
  url          = {https://doi.org/10.48550/arXiv.2310.03714},
  doi          = {10.48550/ARXIV.2310.03714},
  eprinttype    = {arXiv},
  eprint       = {2310.03714},
  bibsource    = {dblp computer science bibliography, https://dblp.org}
}

@article{kranti2024towards,
  title={Towards no-code programming of cobots: Experiments with code synthesis by large code models for conversational programming},
  author={Kranti, Chalamalasetti and Hakimov, Sherzod and Schlangen, David},
  journal={arXiv preprint arXiv:2409.11041},
  year={2024}
}

@inproceedings{colgate1996cobots,
  title={Cobots: Robots for collaboration with human operators},
  author={Colgate, J Edward and Wannasuphoprasit, Witaya and Peshkin, Michael A},
  booktitle={ASME international mechanical engineering congress and exposition},
  volume={15281},
  pages={433--439},
  year={1996},
  organization={American Society of Mechanical Engineers}
}

@article{george2023task,
  title={Task complexity and the skills dilemma in the programming and control of collaborative robots for manufacturing},
  author={George, Peter and Cheng, Chi-Tsun and Pang, Toh Yen and Neville, Katrina},
  journal={Applied Sciences},
  volume={13},
  number={7},
  pages={4635},
  year={2023},
  publisher={MDPI}
}

@inproceedings{DBLP:conf/hri/WhiteM24,
  author       = {Nathan Thomas White and
                  Bilge Mutlu},
  editor       = {Dan Grollman and
                  Elizabeth Broadbent and
                  Wendy Ju and
                  Harold Soh and
                  Tom Williams},
  title        = {End User Interfaces for Human-Robot Collaboration},
  booktitle    = {Companion of the 2024 {ACM/IEEE} International Conference on Human-Robot
                  Interaction, {HRI} 2024, Boulder, CO, USA, March 11-15, 2024},
  pages        = {157--159},
  publisher    = {{ACM}},
  year         = {2024},
  url          = {https://doi.org/10.1145/3610978.3638362},
  doi          = {10.1145/3610978.3638362},
  bibsource    = {dblp computer science bibliography, https://dblp.org}
}

@inproceedings{DBLP:conf/corl/IchterBCFHHHIIJ22,
  author       = {Brian Ichter and
                  Anthony Brohan and
                  Yevgen Chebotar and
                  Chelsea Finn and
                  Karol Hausman and
                  Alexander Herzog and
                  Daniel Ho and
                  Julian Ibarz and
                  Alex Irpan and
                  Eric Jang and
                  Ryan Julian and
                  Dmitry Kalashnikov and
                  Sergey Levine and
                  Yao Lu and
                  Carolina Parada and
                  Kanishka Rao and
                  Pierre Sermanet and
                  Alexander Toshev and
                  Vincent Vanhoucke and
                  Fei Xia and
                  Ted Xiao and
                  Peng Xu and
                  Mengyuan Yan and
                  Noah Brown and
                  Michael Ahn and
                  Omar Cortes and
                  Nicolas Sievers and
                  Clayton Tan and
                  Sichun Xu and
                  Diego Reyes and
                  Jarek Rettinghouse and
                  Jornell Quiambao and
                  Peter Pastor and
                  Linda Luu and
                  Kuang{-}Huei Lee and
                  Yuheng Kuang and
                  Sally Jesmonth and
                  Nikhil J. Joshi and
                  Kyle Jeffrey and
                  Rosario Jauregui Ruano and
                  Jasmine Hsu and
                  Keerthana Gopalakrishnan and
                  Byron David and
                  Andy Zeng and
                  Chuyuan Kelly Fu},
  editor       = {Karen Liu and
                  Dana Kulic and
                  Jeffrey Ichnowski},
  title        = {Do As {I} Can, Not As {I} Say: Grounding Language in Robotic Affordances},
  booktitle    = {Conference on Robot Learning, CoRL 2022, 14-18 December 2022, Auckland,
                  New Zealand},
  series       = {Proceedings of Machine Learning Research},
  volume       = {205},
  pages        = {287--318},
  publisher    = {{PMLR}},
  year         = {2022},
  url          = {https://proceedings.mlr.press/v205/ichter23a.html},
  bibsource    = {dblp computer science bibliography, https://dblp.org}
}

@inproceedings{DBLP:conf/icra/SinghBMGXTFTG23,
  author       = {Ishika Singh and
                  Valts Blukis and
                  Arsalan Mousavian and
                  Ankit Goyal and
                  Danfei Xu and
                  Jonathan Tremblay and
                  Dieter Fox and
                  Jesse Thomason and
                  Animesh Garg},
  title        = {ProgPrompt: Generating Situated Robot Task Plans using Large Language
                  Models},
  booktitle    = {{IEEE} International Conference on Robotics and Automation, {ICRA}
                  2023, London, UK, May 29 - June 2, 2023},
  pages        = {11523--11530},
  publisher    = {{IEEE}},
  year         = {2023},
  url          = {https://doi.org/10.1109/ICRA48891.2023.10161317},
  doi          = {10.1109/ICRA48891.2023.10161317},
  bibsource    = {dblp computer science bibliography, https://dblp.org}
}

@inproceedings{DBLP:conf/icra/LiangHXXHIFZ23,
  author       = {Jacky Liang and
                  Wenlong Huang and
                  Fei Xia and
                  Peng Xu and
                  Karol Hausman and
                  Brian Ichter and
                  Pete Florence and
                  Andy Zeng},
  title        = {Code as Policies: Language Model Programs for Embodied Control},
  booktitle    = {{IEEE} International Conference on Robotics and Automation, {ICRA}
                  2023, London, UK, May 29 - June 2, 2023},
  pages        = {9493--9500},
  publisher    = {{IEEE}},
  year         = {2023},
  url          = {https://doi.org/10.1109/ICRA48891.2023.10160591},
  doi          = {10.1109/ICRA48891.2023.10160591},
  bibsource    = {dblp computer science bibliography, https://dblp.org}
}

@inproceedings{DBLP:conf/icra/AkellaPCWNWHPP99,
  author       = {Prasad Akella and
                  Michael A. Peshkin and
                  J. Edward Colgate and
                  Wit Wannasuphoprasit and
                  Nidamaluri Nagesh and
                  Jim Wells and
                  Steve Holland and
                  Tom Pearson and
                  Brian Peacock},
  title        = {Cobots for the Automobile Assembly Line},
  booktitle    = {1999 {IEEE} International Conference on Robotics and Automation, Marriott
                  Hotel, Renaissance Center, Detroit, Michigan, USA, May 10-15, 1999,
                  Proceedings},
  pages        = {728--733},
  publisher    = {{IEEE} Robotics and Automation Society},
  year         = {1999},
  url          = {https://doi.org/10.1109/ROBOT.1999.770061},
  doi          = {10.1109/ROBOT.1999.770061},
  bibsource    = {dblp computer science bibliography, https://dblp.org}
}

@article{DBLP:journals/tmlr/WangX0MXZFA24,
  author       = {Guanzhi Wang and
                  Yuqi Xie and
                  Yunfan Jiang and
                  Ajay Mandlekar and
                  Chaowei Xiao and
                  Yuke Zhu and
                  Linxi Fan and
                  Anima Anandkumar},
  title        = {Voyager: An Open-Ended Embodied Agent with Large Language Models},
  journal      = {Trans. Mach. Learn. Res.},
  volume       = {2024},
  year         = {2024},
  url          = {https://openreview.net/forum?id=ehfRiF0R3a},
  bibsource    = {dblp computer science bibliography, https://dblp.org}
}

@inproceedings{DBLP:conf/hri/KimLM24,
  author       = {Callie Y. Kim and
                  Christine P. Lee and
                  Bilge Mutlu},
  editor       = {Dan Grollman and
                  Elizabeth Broadbent and
                  Wendy Ju and
                  Harold Soh and
                  Tom Williams},
  title        = {Understanding Large-Language Model (LLM)-powered Human-Robot Interaction},
  booktitle    = {Proceedings of the 2024 {ACM/IEEE} International Conference on Human-Robot
                  Interaction, {HRI} 2024, Boulder, CO, USA, March 11-15, 2024},
  pages        = {371--380},
  publisher    = {{ACM}},
  year         = {2024},
  url          = {https://doi.org/10.1145/3610977.3634966},
  doi          = {10.1145/3610977.3634966},
  bibsource    = {dblp computer science bibliography, https://dblp.org}
}

@inproceedings{DBLP:conf/icml/HuangAPM22,
  author       = {Wenlong Huang and
                  Pieter Abbeel and
                  Deepak Pathak and
                  Igor Mordatch},
  editor       = {Kamalika Chaudhuri and
                  Stefanie Jegelka and
                  Le Song and
                  Csaba Szepesv{\'{a}}ri and
                  Gang Niu and
                  Sivan Sabato},
  title        = {Language Models as Zero-Shot Planners: Extracting Actionable Knowledge
                  for Embodied Agents},
  booktitle    = {International Conference on Machine Learning, {ICML} 2022, 17-23 July
                  2022, Baltimore, Maryland, {USA}},
  series       = {Proceedings of Machine Learning Research},
  volume       = {162},
  pages        = {9118--9147},
  publisher    = {{PMLR}},
  year         = {2022},
  url          = {https://proceedings.mlr.press/v162/huang22a.html},
  bibsource    = {dblp computer science bibliography, https://dblp.org}
}

@inproceedings{DBLP:conf/icml/DriessXSLCIWTVY23,
  author       = {Danny Driess and
                  Fei Xia and
                  Mehdi S. M. Sajjadi and
                  Corey Lynch and
                  Aakanksha Chowdhery and
                  Brian Ichter and
                  Ayzaan Wahid and
                  Jonathan Tompson and
                  Quan Vuong and
                  Tianhe Yu and
                  Wenlong Huang and
                  Yevgen Chebotar and
                  Pierre Sermanet and
                  Daniel Duckworth and
                  Sergey Levine and
                  Vincent Vanhoucke and
                  Karol Hausman and
                  Marc Toussaint and
                  Klaus Greff and
                  Andy Zeng and
                  Igor Mordatch and
                  Pete Florence},
  editor       = {Andreas Krause and
                  Emma Brunskill and
                  Kyunghyun Cho and
                  Barbara Engelhardt and
                  Sivan Sabato and
                  Jonathan Scarlett},
  title        = {PaLM-E: An Embodied Multimodal Language Model},
  booktitle    = {International Conference on Machine Learning, {ICML} 2023, 23-29 July
                  2023, Honolulu, Hawaii, {USA}},
  series       = {Proceedings of Machine Learning Research},
  volume       = {202},
  pages        = {8469--8488},
  publisher    = {{PMLR}},
  year         = {2023},
  url          = {https://proceedings.mlr.press/v202/driess23a.html},
  bibsource    = {dblp computer science bibliography, https://dblp.org}
}

@inproceedings{DBLP:conf/aaai/PadmakumarTSLNG22,
  author       = {Aishwarya Padmakumar and
                  Jesse Thomason and
                  Ayush Shrivastava and
                  Patrick Lange and
                  Anjali Narayan{-}Chen and
                  Spandana Gella and
                  Robinson Piramuthu and
                  G{\"{o}}khan T{\"{u}}r and
                  Dilek Hakkani{-}T{\"{u}}r},
  title        = {TEACh: Task-Driven Embodied Agents That Chat},
  booktitle    = {Thirty-Sixth {AAAI} Conference on Artificial Intelligence, {AAAI}
                  2022, Thirty-Fourth Conference on Innovative Applications of Artificial
                  Intelligence, {IAAI} 2022, The Twelveth Symposium on Educational Advances
                  in Artificial Intelligence, {EAAI} 2022 Virtual Event, February 22
                  - March 1, 2022},
  pages        = {2017--2025},
  publisher    = {{AAAI} Press},
  year         = {2022},
  url          = {https://doi.org/10.1609/aaai.v36i2.20097},
  doi          = {10.1609/AAAI.V36I2.20097},
  bibsource    = {dblp computer science bibliography, https://dblp.org}
}

@article{DBLP:journals/corr/abs-1910-11471,
  author       = {K. M. Tahsin Hassan Rahit and
                  Rashidul Hasan Nabil and
                  Md Hasibul Huq},
  title        = {Machine Translation from Natural Language to Code using Long-Short
                  Term Memory},
  journal      = {CoRR},
  volume       = {abs/1910.11471},
  year         = {2019},
  url          = {http://arxiv.org/abs/1910.11471},
  eprinttype    = {arXiv},
  eprint       = {1910.11471},
  bibsource    = {dblp computer science bibliography, https://dblp.org}
}

@article{DBLP:journals/ijst/Desilets01,
  author       = {Alain D{\'{e}}silets},
  title        = {VoiceGrip: {A} Tool for Programming-by-Voice},
  journal      = {Int. J. Speech Technol.},
  volume       = {4},
  number       = {2},
  pages        = {103--116},
  year         = {2001},
  url          = {https://doi.org/10.1023/A\%3A1011323308477},
  doi          = {10.1023/A\%3A1011323308477},
  bibsource    = {dblp computer science bibliography, https://dblp.org}
}

@book{cypher1993watch,
  title={Watch what I do: programming by demonstration},
  author={Cypher, Allen and Halbert, Daniel Conrad},
  year={1993},
  publisher={MIT press}
}

@article{DBLP:journals/ras/LauriaBKK02,
  author       = {Stanislao Lauria and
                  Guido Bugmann and
                  Theocharis Kyriacou and
                  Ewan Klein},
  title        = {Mobile robot programming using natural language},
  journal      = {Robotics Auton. Syst.},
  volume       = {38},
  number       = {3-4},
  pages        = {171--181},
  year         = {2002},
  url          = {https://doi.org/10.1016/S0921-8890(02)00166-5},
  doi          = {10.1016/S0921-8890(02)00166-5},
  bibsource    = {dblp computer science bibliography, https://dblp.org}
}

@inproceedings{DBLP:conf/vl/GeDSLHS24,
  author       = {Yate Ge and
                  Yi Dai and
                  Run Shan and
                  Kechun Li and
                  Yuanda Hu and
                  Xiaohua Sun},
  title        = {Cocobo: Exploring Large Language Models as the Engine for End-User
                  Robot Programming},
  booktitle    = {2024 {IEEE} Symposium on Visual Languages and Human-Centric Computing
                  (VL/HCC), Liverpool, UK, September 2-6, 2024},
  pages        = {89--95},
  publisher    = {{IEEE}},
  year         = {2024},
  url          = {https://doi.org/10.1109/VL/HCC60511.2024.00020},
  doi          = {10.1109/VL/HCC60511.2024.00020},
  bibsource    = {dblp computer science bibliography, https://dblp.org}
}

@inproceedings{10.1109/ICRA.2019.8794287,
author = {Thomason, Jesse and Padmakumar, Aishwarya and Sinapov, Jivko and Walker, Nick and Jiang, Yuqian and Yedidsion, Harel and Hart, Justin and Stone, Peter and Mooney, Raymond J.},
title = {Improving Grounded Natural Language Understanding through Human-Robot Dialog},
year = {2019},
publisher = {IEEE Press},
url = {https://doi.org/10.1109/ICRA.2019.8794287},
doi = {10.1109/ICRA.2019.8794287},
booktitle = {2019 International Conference on Robotics and Automation (ICRA)},
pages = {6934–6941},
numpages = {8},
location = {Montreal, QC, Canada}
}

@inproceedings{chalamalasetti-etal-2023-clembench,
    title = "clembench: Using Game Play to Evaluate Chat-Optimized Language Models as Conversational Agents",
    author = {Chalamalasetti, Kranti  and
      G{\"o}tze, Jana  and
      Hakimov, Sherzod  and
      Madureira, Brielen  and
      Sadler, Philipp  and
      Schlangen, David},
    editor = "Bouamor, Houda  and
      Pino, Juan  and
      Bali, Kalika",
    booktitle = "Proceedings of the 2023 Conference on Empirical Methods in Natural Language Processing",
    month = dec,
    year = "2023",
    address = "Singapore",
    publisher = "Association for Computational Linguistics",
    url = "https://aclanthology.org/2023.emnlp-main.689/",
    doi = "10.18653/v1/2023.emnlp-main.689",
    pages = "11174--11219"
}

@INPROCEEDINGS{11091799,
  author={Tapia, Endika and Lopez-Novoa, Unai and Sastoque-Pinilla, Leonardo and López-de-Lacalle, Luis Norberto},
  booktitle={2025 10th International Conference on Smart and Sustainable Technologies (SpliTech)}, 
  title={An Evaluation of Large Language Models for Code Optimization}, 
  year={2025},
  volume={},
  number={},
  pages={1-5},
  doi={10.23919/SpliTech65624.2025.11091799}}

@article{DBLP:journals/corr/abs-2309-03409,
  author       = {Chengrun Yang and
                  Xuezhi Wang and
                  Yifeng Lu and
                  Hanxiao Liu and
                  Quoc V. Le and
                  Denny Zhou and
                  Xinyun Chen},
  title        = {Large Language Models as Optimizers},
  journal      = {CoRR},
  volume       = {abs/2309.03409},
  year         = {2023},
  url          = {https://doi.org/10.48550/arXiv.2309.03409},
  doi          = {10.48550/ARXIV.2309.03409},
  eprinttype    = {arXiv},
  eprint       = {2309.03409},
  bibsource    = {dblp computer science bibliography, https://dblp.org}
}

@INPROCEEDINGS{11029992,
  author={Cynthia, Shamse Tasnim},
  booktitle={2025 IEEE/ACM 4th International Conference on AI Engineering – Software Engineering for AI (CAIN)}, 
  title={Identification and Optimization of Redundant Code Using Large Language Models}, 
  year={2025},
  volume={},
  number={},
  pages={261-263},
  doi={10.1109/CAIN66642.2025.00042}}

@article{robotics11010009,
author = {Giberti, Hermes and Abbattista, Tommaso and Carnevale, Marco and Giagu, Luca and Cristini, Fabio},
title = {A Methodology for Flexible Implementation of Collaborative Robots in Smart Manufacturing Systems},
journal = {Robotics},
volume = {11},
year = {2022},
number = {1},
article-number = {9},
url = {https://www.mdpi.com/2218-6581/11/1/9},
issn = {2218-6581},
doi = {10.3390/robotics11010009}
}

@article{10.1145/3466819,
author = {Ajaykumar, Gopika and Steele, Maureen and Huang, Chien-Ming},
title = {A Survey on End-User Robot Programming},
year = {2021},
issue_date = {November 2022},
publisher = {Association for Computing Machinery},
address = {New York, NY, USA},
volume = {54},
number = {8},
issn = {0360-0300},
url = {https://doi.org/10.1145/3466819},
doi = {10.1145/3466819},
journal = {ACM Comput. Surv.},
month = oct,
articleno = {164},
numpages = {36}
}

@article{DBLP:journals/ei/AkkaladeviPPIH19,
  author       = {Sharath Chandra Akkaladevi and
                  Andreas Pichler and
                  Matthias Plasch and
                  Markus Ikeda and
                  Michael Hofmann},
  title        = {Skill-based programming of complex robotic assembly tasks for industrial
                  application},
  journal      = {Elektrotech. Informationstechnik},
  volume       = {136},
  number       = {7},
  pages        = {326--333},
  year         = {2019},
  url          = {https://doi.org/10.1007/s00502-019-00741-4},
  doi          = {10.1007/S00502-019-00741-4},
  bibsource    = {dblp computer science bibliography, https://dblp.org}
}

@INPROCEEDINGS{9800018,
  author={Dmytriyev, Yevheniy and Carnevale, Marco and Giberti, Hermes and Todeschini, Giovanni},
  booktitle={2022 International Congress on Human-Computer Interaction, Optimization and Robotic Applications (HORA)}, 
  title={On cobot programming in industrial tasks: a test case}, 
  year={2022},
  volume={},
  number={},
  pages={1-9},
  doi={10.1109/HORA55278.2022.9800018}}

@inproceedings{DBLP:conf/hri/AyalewWLUS25,
  author       = {Tewodros W. Ayalew and
                  Jennifer Wang and
                  Michael L. Littman and
                  Blase Ur and
                  Sarah Sebo},
  title        = {Enabling End Users to Program Robots Using Reinforcement Learning},
  booktitle    = {20th {ACM/IEEE} International Conference on Human-Robot Interaction,
                  {HRI} 2025, Melbourne, Australia, March 4-6, 2025},
  pages        = {767--777},
  publisher    = {{IEEE}},
  year         = {2025},
  url          = {https://doi.org/10.1109/HRI61500.2025.10974035},
  doi          = {10.1109/HRI61500.2025.10974035},
  bibsource    = {dblp computer science bibliography, https://dblp.org}
}

@article{barravecchia2023general,
  title={A general cost model to assess the implementation of collaborative robots in assembly processes},
  author={Barravecchia, Federico and Mastrogiacomo, Luca and Franceschini, Fiorenzo},
  journal={The International Journal of Advanced Manufacturing Technology},
  volume={125},
  number={11},
  pages={5247--5266},
  year={2023},
  publisher={Springer}
}

@inproceedings{kane-etal-2022-system,
    title = "A System For Robot Concept Learning Through Situated Dialogue",
    author = "Kane, Benjamin  and
      Gervits, Felix  and
      Scheutz, Matthias  and
      Marge, Matthew",
    editor = "Lemon, Oliver  and
      Hakkani-Tur, Dilek  and
      Li, Junyi Jessy  and
      Ashrafzadeh, Arash  and
      Garcia, Daniel Hern{\'a}ndez  and
      Alikhani, Malihe  and
      Vandyke, David  and
      Du{\v{s}}ek, Ond{\v{r}}ej",
    booktitle = "Proceedings of the 23rd Annual Meeting of the Special Interest Group on Discourse and Dialogue",
    month = sep,
    year = "2022",
    address = "Edinburgh, UK",
    publisher = "Association for Computational Linguistics",
    url = "https://aclanthology.org/2022.sigdial-1.64/",
    doi = "10.18653/v1/2022.sigdial-1.64",
    pages = "659--662"
}

@article{GOROSTIZA20111102,
title = {End-user programming of a social robot by dialog},
journal = {Robotics and Autonomous Systems},
volume = {59},
number = {12},
pages = {1102-1114},
year = {2011},
issn = {0921-8890},
doi = {https://doi.org/10.1016/j.robot.2011.07.009},
url = {https://www.sciencedirect.com/science/article/pii/S092188901100131X},
author = {Javi F. Gorostiza and Miguel A. Salichs}
}

@inproceedings{10.1145/1180995.1181009,
author = {Foster, Mary Ellen and By, Tomas and Rickert, Markus and Knoll, Alois},
title = {Human-Robot dialogue for joint construction tasks},
year = {2006},
isbn = {159593541X},
publisher = {Association for Computing Machinery},
address = {New York, NY, USA},
url = {https://doi.org/10.1145/1180995.1181009},
doi = {10.1145/1180995.1181009},
booktitle = {Proceedings of the 8th International Conference on Multimodal Interfaces},
pages = {68–71},
numpages = {4},
location = {Banff, Alberta, Canada},
series = {ICMI '06}
}

@article{10.1007/s10579-024-09784-2,
author = {Bonial, Claire and Lukin, Stephanie M. and Abrams, Mitchell and Baker, Anthony and Donatelli, Lucia and Foots, Ashley and Hayes, Cory J. and Henry, Cassidy and Hudson, Taylor and Marge, Matthew and Pollard, Kimberly A. and Artstein, Ron and Traum, David and Voss, Clare R.},
title = {Human–robot dialogue annotation for multi-modal common ground: Human–robot dialogue annotation...},
year = {2024},
issue_date = {Jun 2025},
publisher = {Springer-Verlag},
address = {Berlin, Heidelberg},
volume = {59},
number = {2},
issn = {1574-020X},
url = {https://doi.org/10.1007/s10579-024-09784-2},
doi = {10.1007/s10579-024-09784-2},
journal = {Lang. Resour. Eval.},
month = nov,
pages = {1525–1575},
numpages = {51}
}

@inproceedings{DBLP:conf/slt/KaziLZHT24,
  author       = {Taaha Kazi and
                  Ruiliang Lyu and
                  Sizhe Zhou and
                  Dilek Hakkani{-}T{\"{u}}r and
                  Gokhan Tur},
  title        = {Large Language Models as User-Agents For Evaluating Task-Oriented-Dialogue
                  Systems},
  booktitle    = {{IEEE} Spoken Language Technology Workshop, {SLT} 2024, Macao, December
                  2-5, 2024},
  pages        = {913--920},
  publisher    = {{IEEE}},
  year         = {2024},
  url          = {https://doi.org/10.1109/SLT61566.2024.10832298},
  doi          = {10.1109/SLT61566.2024.10832298},
  bibsource    = {dblp computer science bibliography, https://dblp.org}
}

@article{https://doi.org/10.1609/aimag.v37i4.2684,
author = {Chai, Joyce Y. and Fang, Rui and Liu, Changsong and She, Lanbo},
title = {Collaborative Language Grounding Toward Situated Human-Robot Dialogue},
journal = {AI Magazine},
volume = {37},
number = {4},
pages = {32-45},
doi = {https://doi.org/10.1609/aimag.v37i4.2684},
url = {https://onlinelibrary.wiley.com/doi/abs/10.1609/aimag.v37i4.2684},
eprint = {https://onlinelibrary.wiley.com/doi/pdf/10.1609/aimag.v37i4.2684},
year = {2016}
}

@inproceedings{10.5555/3304415.3304417,
author = {Chai, Joyce Y. and Gao, Qiaozi and She, Lanbo and Yang, Shaohua and Saba-Sadiya, Sari and Xu, Guangyue},
title = {Language to action: towards interactive task learning with physical agents},
year = {2018},
isbn = {9780999241127},
publisher = {AAAI Press},
booktitle = {Proceedings of the 27th International Joint Conference on Artificial Intelligence},
pages = {2–9},
numpages = {8},
location = {Stockholm, Sweden},
series = {IJCAI'18}
}

@inproceedings{chiyah-garcia-etal-2024-repairs,
    title = "Repairs in a Block World: A New Benchmark for Handling User Corrections with Multi-Modal Language Models",
    author = "Chiyah-Garcia, Javier  and
      Suglia, Alessandro  and
      Eshghi, Arash",
    editor = "Al-Onaizan, Yaser  and
      Bansal, Mohit  and
      Chen, Yun-Nung",
    booktitle = "Proceedings of the 2024 Conference on Empirical Methods in Natural Language Processing",
    month = nov,
    year = "2024",
    address = "Miami, Florida, USA",
    publisher = "Association for Computational Linguistics",
    url = "https://aclanthology.org/2024.emnlp-main.643/",
    doi = "10.18653/v1/2024.emnlp-main.643",
    pages = "11523--11542"
}

@ARTICLE{5391077,
  author={Heidorn, G. E.},
  journal={IBM Journal of Research and Development}, 
  title={Automatic Programming Through Natural Language Dialogue: A Survey}, 
  year={1976},
  volume={20},
  number={4},
  pages={302-313},
  doi={10.1147/rd.204.0302}}

@article{LEGGETT1984493,
title = {An empirical investigation of voice as an input modality for computer programming},
journal = {International Journal of Man-Machine Studies},
volume = {21},
number = {6},
pages = {493-520},
year = {1984},
issn = {0020-7373},
doi = {https://doi.org/10.1016/S0020-7373(84)80057-7},
url = {https://www.sciencedirect.com/science/article/pii/S0020737384800577},
author = {John Leggett and Glen Williams}
}

@article{DBLP:journals/corr/abs-2107-03374,
  author       = {Mark Chen and
                  Jerry Tworek and
                  Heewoo Jun and
                  Qiming Yuan and
                  Henrique Pond{\'{e}} de Oliveira Pinto and
                  Jared Kaplan and
                  Harri Edwards and
                  Yuri Burda and
                  Nicholas Joseph and
                  Greg Brockman and
                  Alex Ray and
                  Raul Puri and
                  Gretchen Krueger and
                  Michael Petrov and
                  Heidy Khlaaf and
                  Girish Sastry and
                  Pamela Mishkin and
                  Brooke Chan and
                  Scott Gray and
                  Nick Ryder and
                  Mikhail Pavlov and
                  Alethea Power and
                  Lukasz Kaiser and
                  Mohammad Bavarian and
                  Clemens Winter and
                  Philippe Tillet and
                  Felipe Petroski Such and
                  Dave Cummings and
                  Matthias Plappert and
                  Fotios Chantzis and
                  Elizabeth Barnes and
                  Ariel Herbert{-}Voss and
                  William Hebgen Guss and
                  Alex Nichol and
                  Alex Paino and
                  Nikolas Tezak and
                  Jie Tang and
                  Igor Babuschkin and
                  Suchir Balaji and
                  Shantanu Jain and
                  William Saunders and
                  Christopher Hesse and
                  Andrew N. Carr and
                  Jan Leike and
                  Joshua Achiam and
                  Vedant Misra and
                  Evan Morikawa and
                  Alec Radford and
                  Matthew Knight and
                  Miles Brundage and
                  Mira Murati and
                  Katie Mayer and
                  Peter Welinder and
                  Bob McGrew and
                  Dario Amodei and
                  Sam McCandlish and
                  Ilya Sutskever and
                  Wojciech Zaremba},
  title        = {Evaluating Large Language Models Trained on Code},
  journal      = {CoRR},
  volume       = {abs/2107.03374},
  year         = {2021},
  url          = {https://arxiv.org/abs/2107.03374},
  eprinttype   = {arXiv},
  eprint       = {2107.03374},
  bibsource    = {dblp computer science bibliography, https://dblp.org}
}

@article{DBLP:journals/corr/abs-2108-07732,
  author       = {Jacob Austin and
                  Augustus Odena and
                  Maxwell I. Nye and
                  Maarten Bosma and
                  Henryk Michalewski and
                  David Dohan and
                  Ellen Jiang and
                  Carrie J. Cai and
                  Michael Terry and
                  Quoc V. Le and
                  Charles Sutton},
  title        = {Program Synthesis with Large Language Models},
  journal      = {CoRR},
  volume       = {abs/2108.07732},
  year         = {2021},
  url          = {https://arxiv.org/abs/2108.07732},
  eprinttype   = {arXiv},
  eprint       = {2108.07732},
  bibsource    = {dblp computer science bibliography, https://dblp.org}
}

@article{DBLP:journals/tosem/HouZLYWLLLGW24,
  author       = {Xinyi Hou and
                  Yanjie Zhao and
                  Yue Liu and
                  Zhou Yang and
                  Kailong Wang and
                  Li Li and
                  Xiapu Luo and
                  David Lo and
                  John Grundy and
                  Haoyu Wang},
  title        = {Large Language Models for Software Engineering: {A} Systematic Literature
                  Review},
  journal      = {{ACM} Trans. Softw. Eng. Methodol.},
  volume       = {33},
  number       = {8},
  pages        = {220:1--220:79},
  year         = {2024},
  url          = {https://doi.org/10.1145/3695988},
  doi          = {10.1145/3695988},
  bibsource    = {dblp computer science bibliography, https://dblp.org}
}

@inproceedings{DBLP:conf/iclr/NijkampPHTWZSX23,
  author       = {Erik Nijkamp and
                  Bo Pang and
                  Hiroaki Hayashi and
                  Lifu Tu and
                  Huan Wang and
                  Yingbo Zhou and
                  Silvio Savarese and
                  Caiming Xiong},
  title        = {CodeGen: An Open Large Language Model for Code with Multi-Turn Program
                  Synthesis},
  booktitle    = {The Eleventh International Conference on Learning Representations,
                  {ICLR} 2023, Kigali, Rwanda, May 1-5, 2023},
  publisher    = {OpenReview.net},
  year         = {2023},
  url          = {https://openreview.net/forum?id=iaYcJKpY2B\_},
  bibsource    = {dblp computer science bibliography, https://dblp.org}
}

@inproceedings{DBLP:conf/irc/TobergC23,
  author       = {Jan{-}Philipp T{\"{o}}berg and
                  Philipp Cimiano},
  title        = {Generation of Robot Manipulation Plans Using Generative Large Language
                  Models},
  booktitle    = {Seventh {IEEE} International Conference on Robotic Computing, {IRC}
                  2023, Laguna Hills, CA, USA, December 11-13, 2023},
  pages        = {190--197},
  publisher    = {{IEEE}},
  year         = {2023},
  url          = {https://doi.org/10.1109/IRC59093.2023.00039},
  doi          = {10.1109/IRC59093.2023.00039},
  bibsource    = {dblp computer science bibliography, https://dblp.org}
}

@InProceedings{pmlr-v235-mu24a,
  title = 	 {{R}obo{C}ode{X}: Multimodal Code Generation for Robotic Behavior Synthesis},
  author =       {Mu, Yao and Chen, Junting and Zhang, Qing-Long and Chen, Shoufa and Yu, Qiaojun and Ge, Chongjian and Chen, Runjian and Liang, Zhixuan and Hu, Mengkang and Tao, Chaofan and Sun, Peize and Yu, Haibao and Yang, Chao and Shao, Wenqi and Wang, Wenhai and Dai, Jifeng and Qiao, Yu and Ding, Mingyu and Luo, Ping},
  booktitle = 	 {Proceedings of the 41st International Conference on Machine Learning},
  pages = 	 {36434--36454},
  year = 	 {2024},
  editor = 	 {Salakhutdinov, Ruslan and Kolter, Zico and Heller, Katherine and Weller, Adrian and Oliver, Nuria and Scarlett, Jonathan and Berkenkamp, Felix},
  volume = 	 {235},
  series = 	 {Proceedings of Machine Learning Research},
  month = 	 {21--27 Jul},
  publisher =    {PMLR},
  url = 	 {https://proceedings.mlr.press/v235/mu24a.html}
}

@inproceedings{DBLP:conf/icra/MacalusoCC24,
  author       = {Annabella Macaluso and
                  Nicholas Cote and
                  Sachin Chitta},
  title        = {Toward Automated Programming for Robotic Assembly Using ChatGPT},
  booktitle    = {{IEEE} International Conference on Robotics and Automation, {ICRA}
                  2024, Yokohama, Japan, May 13-17, 2024},
  pages        = {17687--17693},
  publisher    = {{IEEE}},
  year         = {2024},
  url          = {https://doi.org/10.1109/ICRA57147.2024.10610554},
  doi          = {10.1109/ICRA57147.2024.10610554},
  bibsource    = {dblp computer science bibliography, https://dblp.org}
}

@article{DBLP:journals/corr/abs-2405-20179,
  author       = {Zichao Hu and
                  Junyi Jessy Li and
                  Arjun Guha and
                  Joydeep Biswas},
  title        = {Robo-Instruct: Simulator-Augmented Instruction Alignment For Finetuning
                  CodeLLMs},
  journal      = {CoRR},
  volume       = {abs/2405.20179},
  year         = {2024},
  url          = {https://doi.org/10.48550/arXiv.2405.20179},
  doi          = {10.48550/ARXIV.2405.20179},
  eprinttype   = {arXiv},
  eprint       = {2405.20179},
  bibsource    = {dblp computer science bibliography, https://dblp.org}
}

@article{DBLP:journals/ijrr/KonidarisKGB12,
  author       = {George Dimitri Konidaris and
                  Scott Kuindersma and
                  Roderic A. Grupen and
                  Andrew G. Barto},
  title        = {Robot learning from demonstration by constructing skill trees},
  journal      = {Int. J. Robotics Res.},
  volume       = {31},
  number       = {3},
  pages        = {360--375},
  year         = {2012},
  url          = {https://doi.org/10.1177/0278364911428653},
  doi          = {10.1177/0278364911428653},
  bibsource    = {dblp computer science bibliography, https://dblp.org}
}

@article{DBLP:journals/ijrr/LioutikovNMP17,
  author       = {Rudolf Lioutikov and
                  Gerhard Neumann and
                  Guilherme Maeda and
                  Jan Peters},
  title        = {Learning movement primitive libraries through probabilistic segmentation},
  journal      = {Int. J. Robotics Res.},
  volume       = {36},
  number       = {8},
  pages        = {879--894},
  year         = {2017},
  url          = {https://doi.org/10.1177/0278364917713116},
  doi          = {10.1177/0278364917713116},
  bibsource    = {dblp computer science bibliography, https://dblp.org}
}

@inproceedings{DBLP:conf/icra/Tziafas024,
  author       = {Georgios Tziafas and
                  Hamidreza Kasaei},
  title        = {Lifelong Robot Library Learning: Bootstrapping Composable and Generalizable
                  Skills for Embodied Control with Language Models},
  booktitle    = {{IEEE} International Conference on Robotics and Automation, {ICRA}
                  2024, Yokohama, Japan, May 13-17, 2024},
  pages        = {515--522},
  publisher    = {{IEEE}},
  year         = {2024},
  url          = {https://doi.org/10.1109/ICRA57147.2024.10611448},
  doi          = {10.1109/ICRA57147.2024.10611448},
  bibsource    = {dblp computer science bibliography, https://dblp.org}
}

@ARTICLE{1456949,
  author={Lozano-Perez, T.},
  journal={Proceedings of the IEEE}, 
  title={Robot programming}, 
  year={1983},
  volume={71},
  number={7},
  pages={821-841},
  doi={10.1109/PROC.1983.12681}}

@inproceedings{biggs2003survey,
  title={A survey of robot programming systems},
  author={Biggs, Geoffrey and MacDonald, Bruce and others},
  booktitle={Proceedings of the Australasian conference on robotics and automation},
  volume={1},
  pages={1--3},
  year={2003}
}

@article{russmann2015industry,
  title={Industry 4.0: The future of productivity and growth in manufacturing industries},
  author={R{\"u}{\ss}mann, Michael and Lorenz, Markus and Gerbert, Philipp and Waldner, Manuela and Justus, Jan and Engel, Pascal and Harnisch, Michael},
  journal={Boston consulting group},
  volume={9},
  number={1},
  pages={54--89},
  year={2015},
  publisher={Boston, MA, USA:}
}

@ARTICLE{9493209,
  author={Rodríguez-Guerra, Diego and Sorrosal, Gorka and Cabanes, Itziar and Calleja, Carlos},
  journal={IEEE Access}, 
  title={Human-Robot Interaction Review: Challenges and Solutions for Modern Industrial Environments}, 
  year={2021},
  volume={9},
  number={},
  pages={108557-108578},
  doi={10.1109/ACCESS.2021.3099287}}

@article{segura2025work,
  title={Work roles in human--robot collaborative systems: Effects on cognitive ergonomics for the manufacturing industry},
  author={Segura, Pablo and Lobato-Calleros, Odette and Soria-Arguello, Isidro and Hern{\'a}ndez-Mart{\'\i}nez, Eduardo Gamaliel},
  journal={Applied Sciences},
  volume={15},
  number={2},
  pages={744},
  year={2025},
  publisher={MDPI}
}

@article{xu2018industry,
  title={Industry 4.0: state of the art and future trends},
  author={Xu, Li Da and Xu, Eric L and Li, Ling},
  journal={International journal of production research},
  volume={56},
  number={8},
  pages={2941--2962},
  year={2018},
  publisher={Taylor \& Francis}
}

@INPROCEEDINGS{606888,
  author={Wannasuphoprasit, W. and Gillespie, R.B. and Colgate, J.E. and Peshkin, M.A.},
  booktitle={Proceedings of International Conference on Robotics and Automation}, 
  title={Cobot control}, 
  year={1997},
  volume={4},
  number={},
  pages={3571-3576 vol.4},
  doi={10.1109/ROBOT.1997.606888}}

@INPROCEEDINGS{981926,
  author={Stopp, A. and Horstmann, S. and Kristensen, S. and Lohnert, F.},
  booktitle={Proceedings 10th IEEE International Workshop on Robot and Human Interactive Communication. ROMAN 2001 (Cat. No.01TH8591)}, 
  title={Towards interactive learning for manufacturing assistants}, 
  year={2001},
  volume={},
  number={},
  pages={338-342},
  doi={10.1109/ROMAN.2001.981926}}

@INPROCEEDINGS{1045667,
  author={Estable, S. and Ahms, I. and Backhaus, H.G. and El Zubi, O. and Muenstermann, R.},
  booktitle={Proceedings. 11th IEEE International Workshop on Robot and Human Interactive Communication}, 
  title={Intuitive teaching and surveillance for production assistants}, 
  year={2002},
  volume={},
  number={},
  pages={474-481},
  doi={10.1109/ROMAN.2002.1045667}}

@INPROCEEDINGS{7745145,
  author={Andersen, Rasmus S. and Madsen, Ole and Moeslund, Thomas B. and Amor, Heni Ben},
  booktitle={2016 25th IEEE International Symposium on Robot and Human Interactive Communication (RO-MAN)}, 
  title={Projecting robot intentions into human environments}, 
  year={2016},
  volume={},
  number={},
  pages={294-301},
  doi={10.1109/ROMAN.2016.7745145}}

@InProceedings{10.1007/978-3-030-28619-4_26,
author="Rosen, Eric
and Whitney, David
and Phillips, Elizabeth
and Chien, Gary
and Tompkin, James
and Konidaris, George
and Tellex, Stefanie",
editor="Amato, Nancy M.
and Hager, Greg
and Thomas, Shawna
and Torres-Torriti, Miguel",
title="Communicating Robot Arm Motion Intent Through Mixed Reality Head-Mounted Displays",
booktitle="Robotics Research",
year="2020",
publisher="Springer International Publishing",
address="Cham",
pages="301--316",
isbn="978-3-030-28619-4"
}

@INPROCEEDINGS{9889345,
  author={Schoen, Andrew and White, Nathan and Henrichs, Curt and Siebert-Evenstone, Amanda and Shaffer, David and Mutlu, Bilge},
  booktitle={2022 17th ACM/IEEE International Conference on Human-Robot Interaction (HRI)}, 
  title={CoFrame: A System for Training Novice Cobot Programmers}, 
  year={2022},
  volume={},
  number={},
  pages={185-194},
  doi={10.1109/HRI53351.2022.9889345}}

@article{ELZAATARI2019162,
title = {Cobot programming for collaborative industrial tasks: An overview},
journal = {Robotics and Autonomous Systems},
volume = {116},
pages = {162-180},
year = {2019},
issn = {0921-8890},
doi = {https://doi.org/10.1016/j.robot.2019.03.003},
url = {https://www.sciencedirect.com/science/article/pii/S092188901830602X},
author = {Shirine {El Zaatari} and Mohamed Marei and Weidong Li and Zahid Usman}
}

@ARTICLE{9320477,
  author={Kuhail, Mohammad Amin and Farooq, Shahbano and Hammad, Rawad and Bahja, Mohammed},
  journal={IEEE Access}, 
  title={Characterizing Visual Programming Approaches for End-User Developers: A Systematic Review}, 
  year={2021},
  volume={9},
  number={},
  pages={14181-14202},
  doi={10.1109/ACCESS.2021.3051043}}

@article{AJAYKUMAR2021103845,
title = {Designing user-centric programming aids for kinesthetic teaching of collaborative robots},
journal = {Robotics and Autonomous Systems},
volume = {145},
pages = {103845},
year = {2021},
issn = {0921-8890},
doi = {https://doi.org/10.1016/j.robot.2021.103845},
url = {https://www.sciencedirect.com/science/article/pii/S0921889021001305},
author = {Gopika Ajaykumar and Maia Stiber and Chien-Ming Huang}
}

@inproceedings{beschi2019capirci,
  title={CAPIRCI: a multi-modal system for collaborative robot programming},
  author={Beschi, Sara and Fogli, Daniela and Tampalini, Fabio},
  booktitle={International symposium on end user development},
  pages={51--66},
  year={2019},
  organization={Springer}
}

@article{tsarouchi2016human,
  title={Human--robot interaction review and challenges on task planning and programming},
  author={Tsarouchi, Panagiota and Makris, Sotiris and Chryssolouris, George},
  journal={International Journal of Computer Integrated Manufacturing},
  volume={29},
  number={8},
  pages={916--931},
  year={2016},
  publisher={Taylor \& Francis}
}

@article{ZHANG2023100131,
title = {Large language models for human–robot interaction: A review},
journal = {Biomimetic Intelligence and Robotics},
volume = {3},
number = {4},
pages = {100131},
year = {2023},
issn = {2667-3797},
doi = {https://doi.org/10.1016/j.birob.2023.100131},
url = {https://www.sciencedirect.com/science/article/pii/S2667379723000451},
author = {Ceng Zhang and Junxin Chen and Jiatong Li and Yanhong Peng and Zebing Mao}
}

@article{ionescu2023programming,
  title={Programming cobots by voice: a pragmatic, web-based approach},
  author={Ionescu, Tudor B and Schlund, Sebastian},
  journal={International Journal of Computer Integrated Manufacturing},
  volume={36},
  number={1},
  pages={86--109},
  year={2023},
  publisher={Taylor \& Francis}
}

@article{FOGLI2022102234,
title = {A hybrid approach to user-oriented programming of collaborative robots},
journal = {Robotics and Computer-Integrated Manufacturing},
volume = {73},
pages = {102234},
year = {2022},
issn = {0736-5845},
doi = {https://doi.org/10.1016/j.rcim.2021.102234},
url = {https://www.sciencedirect.com/science/article/pii/S073658452100106X},
author = {Daniela Fogli and Luigi Gargioni and Giovanni Guida and Fabio Tampalini}
}

@inproceedings{10.1145/3610977.3634969,
author = {Karli, Ulas Berk and Chen, Juo-Tung and Antony, Victor Nikhil and Huang, Chien-Ming},
title = {Alchemist: LLM-Aided End-User Development of Robot Applications},
year = {2024},
isbn = {9798400703225},
publisher = {Association for Computing Machinery},
address = {New York, NY, USA},
url = {https://doi.org/10.1145/3610977.3634969},
doi = {10.1145/3610977.3634969},
booktitle = {Proceedings of the 2024 ACM/IEEE International Conference on Human-Robot Interaction},
pages = {361–370},
numpages = {10},
location = {Boulder, CO, USA},
series = {HRI '24}
}

@ARTICLE{11534298,
  author={Lai, Yuzhi and Yuan, Shenghai and Li, Peizheng and Zhang, Boya and Kiefer, Benjamin and Deng, Tianchen and Zell, Andreas},
  journal={IEEE Transactions on Automation Science and Engineering}, 
  title={FAM-HRI: Foundation-Model Assisted Multimodal Human–Robot Interaction Combining Gaze and Speech}, 
  year={2026},
  volume={23},
  number={},
  pages={10521-10534},
  doi={10.1109/TASE.2026.3695438}}

@article{DBLP:journals/corr/abs-2607-00272,
  author       = {Runyu Lu and
                  Yubo Wu and
                  Ethan Kou and
                  Letian Fu and
                  Wenli Xiao and
                  Ajay Mandlekar and
                  Yinzhen Xu and
                  Guanya Shi and
                  Ken Goldberg and
                  Ang Chen and
                  Mosharaf Chowdhury and
                  Yuke Zhu and
                  Linxi "Jim" Fan and
                  Guanzhi Wang},
  title        = {{ASPIRE:} Agentic /Skills Discovery for Robotics},
  journal      = {CoRR},
  volume       = {abs/2607.00272},
  year         = {2026},
  url          = {https://doi.org/10.48550/arXiv.2607.00272},
  doi          = {10.48550/ARXIV.2607.00272},
  eprinttype   = {arXiv},
  eprint       = {2607.00272},
  bibsource    = {dblp computer science bibliography, https://dblp.org}
}

@InProceedings{pmlr-v306-fu26u,
  title = 	 {{C}a{P}-X: A Framework for Benchmarking and Improving Coding Agents for Robot Manipulation},
  author =       {Fu, Letian and Yu, Justin and El-Refai, Karim and Kou, Ethan and Xue, Haoru and Huang, Huang and Xiao, Wenli and Fei-Fei, Li and Shi, Guanya and Wu, Jiajun and Sastry, S. Shankar and Zhu, Yuke and Goldberg, Ken and Fan, Linxi},
  booktitle = 	 {Proceedings of the 43rd International Conference on Machine Learning},
  pages = 	 {32155--32212},
  year = 	 {2026},
  editor = 	 {Zhang, Tong and Dudik, Miroslav and Jaggi, Martin and Agarwal, Alekh and Li, Sharon and Schuurmans, Dale and Zhu, Jerry and Berkenkamp, Felix and Dong, Hanze and Bietti, Alberto},
  volume = 	 {306},
  series = 	 {Proceedings of Machine Learning Research},
  month = 	 {06--11 Jul},
  publisher =    {PMLR},
  url = 	 {https://proceedings.mlr.press/v306/fu26u.html}
}

@article{hu2026simex,
  title={SimEX: Simulation-Integrated Robotics AutoResearch},
  author={Hu, Jiaheng and Martin-Martin, Roberto and Stone, Peter and Duan, Rocky and Jiang, Zhenyu and Shi, Guanya},
  journal={arXiv preprint arXiv:2609.38982},
  year={2026}
}

@article{AALTONEN20191151,
title = {Experiences and expectations of collaborative robots in industry and academia: barriers and development needs},
journal = {Procedia Manufacturing},
volume = {38},
pages = {1151-1158},
year = {2019},
note = {29th International Conference on Flexible Automation and Intelligent Manufacturing ( FAIM 2019), June 24-28, 2019, Limerick, Ireland, Beyond Industry 4.0: Industrial Advances, Engineering Education and Intelligent Manufacturing},
issn = {2351-9789},
doi = {https://doi.org/10.1016/j.promfg.2020.01.204},
url = {https://www.sciencedirect.com/science/article/pii/S2351978920302055},
author = {Iina Aaltonen and Timo Salmi}
}

@article{bauer2016lightweight,
  title={Lightweight robots in manual assembly--best to start simply},
  author={Bauer, Wilhelm and Bender, Manfred and Braun, Martin and Rally, PETER and Scholtz, Oliver},
  journal={Fraunhofer Institute for Industrial Engineering and Organization IAO, Stuttgart},
  volume={1},
  year={2016}
}

@article{VILLANI2018248,
title = {Survey on human–robot collaboration in industrial settings: Safety, intuitive interfaces and applications},
journal = {Mechatronics},
volume = {55},
pages = {248-266},
year = {2018},
issn = {0957-4158},
doi = {https://doi.org/10.1016/j.mechatronics.2018.02.009},
url = {https://www.sciencedirect.com/science/article/pii/S0957415818300321},
author = {Valeria Villani and Fabio Pini and Francesco Leali and Cristian Secchi}
}

@misc{robots2024cobots,
  title={Why Cobots? All the Benefits of Collaborative Robots},
  author={Robots, Universal},
  year={2024},
  publisher={Retrieved 8/11/2021 from https://www. universal-robots. com/products~…}
}

@inproceedings{10.1145/3610977.3634937,
author = {Sullivan, Dakota and White, Nathan Thomas and Schoen, Andrew and Mutlu, Bilge},
title = {Making Informed Decisions: Supporting Cobot Integration Considering Business and Worker Preferences},
year = {2024},
isbn = {9798400703225},
publisher = {Association for Computing Machinery},
address = {New York, NY, USA},
url = {https://doi.org/10.1145/3610977.3634937},
doi = {10.1145/3610977.3634937},
booktitle = {Proceedings of the 2024 ACM/IEEE International Conference on Human-Robot Interaction},
pages = {706–714},
numpages = {9},
location = {Boulder, CO, USA},
series = {HRI '24}
}

@inproceedings{10.1145/3757279.3785550,
author = {Kim, Callie Y. and White, Nathan Thomas and He, Evan and Sala, Frederic and Mutlu, Bilge},
title = {RoboCritics: Enabling Reliable End-to-End LLM Robot Programming through Expert-Informed Critics},
year = {2026},
isbn = {9798400721281},
publisher = {Association for Computing Machinery},
address = {New York, NY, USA},
url = {https://doi.org/10.1145/3757279.3785550},
doi = {10.1145/3757279.3785550},
pages = {914–923},
numpages = {10},
location = {Edinburgh, Scotland, UK},
series = {HRI '26}
}

@inproceedings{DBLP:conf/nips/BrownMRSKDNSSAA20,
  author       = {Tom B. Brown and
                  Benjamin Mann and
                  Nick Ryder and
                  Melanie Subbiah and
                  Jared Kaplan and
                  et al
},
  editor       = {Hugo Larochelle and
                  Marc'Aurelio Ranzato and
                  Raia Hadsell and
                  Maria{-}Florina Balcan and
                  Hsuan{-}Tien Lin},
  title        = {Language Models are Few-Shot Learners},
  booktitle    = {Advances in Neural Information Processing Systems 33: Annual Conference
                  on Neural Information Processing Systems 2020, NeurIPS 2020, December
                  6-12, 2020, virtual},
  year         = {2020},
  url          = {https://proceedings.neurips.cc/paper/2020/hash/1457c0d6bfcb4967418bfb8ac142f64a-Abstract.html},
  bibsource    = {dblp computer science bibliography, https://dblp.org}
}

@inproceedings{DBLP:conf/nips/Wei0SBIXCLZ22,
  author       = {Jason Wei and
                  Xuezhi Wang and
                  Dale Schuurmans and
                  Maarten Bosma and
                  Brian Ichter and
                  Fei Xia and
                  Ed H. Chi and
                  Quoc V. Le and
                  Denny Zhou},
  editor       = {Sanmi Koyejo and
                  S. Mohamed and
                  A. Agarwal and
                  Danielle Belgrave and
                  K. Cho and
                  A. Oh},
  title        = {Chain-of-Thought Prompting Elicits Reasoning in Large Language Models},
  booktitle    = {Advances in Neural Information Processing Systems 35: Annual Conference
                  on Neural Information Processing Systems 2022, NeurIPS 2022, New Orleans,
                  LA, USA, November 28 - December 9, 2022},
  year         = {2022},
  url          = {http://papers.nips.cc/paper\_files/paper/2022/hash/9d5609613524ecf4f15af0f7b31abca4-Abstract-Conference.html},
  bibsource    = {dblp computer science bibliography, https://dblp.org}
}

@article{DBLP:journals/csur/LiuYFJHN23,
  author       = {Pengfei Liu and
                  Weizhe Yuan and
                  Jinlan Fu and
                  Zhengbao Jiang and
                  Hiroaki Hayashi and
                  Graham Neubig},
  title        = {Pre-train, Prompt, and Predict: {A} Systematic Survey of Prompting
                  Methods in Natural Language Processing},
  journal      = {{ACM} Comput. Surv.},
  volume       = {55},
  number       = {9},
  pages        = {195:1--195:35},
  year         = {2023},
  url          = {https://doi.org/10.1145/3560815},
  doi          = {10.1145/3560815},
  bibsource    = {dblp computer science bibliography, https://dblp.org}
}
\newpage
\section{Appendix}
\label{sec:appendix}
\subsection{Interaction Framework}
\label{subsec:appendix-gamemaster}
As mentioned in Section~\ref{subsec:methodologytaa} of the main content, the TAA interaction framework follows the clem:todd~\citep{kranti-etal-2025-clem} player interaction. The Game Master orchestrates the interaction between both players: the Programmer and the Cobot. Both players are text-based LLMs. In self-play, the same LLM plays the roles of both the Programmer and the Robot, while in cross-play, different models are used as the Programmer and the Robot. While the Programmer is instructed to respond with the follow-up instruction in string format, the Robot is instructed to respond with a JSON object containing the keys \textit{status} and \textit{details}. The status can have the values \textit{code}, \textit{clarification}, or \textit{acknowledgement}. The details will vary depending on the status. If the status is code, the details will contain executable code, or a string if it is clarification or acknowledgement. The Game Master receives the information from each player, strips off the anchors, and routes the response to the next step. In the case of a Robot response, it also validates whether the expected structure is followed, i.e., whether the status and details keys exist and hold only the expected values. If the status is code, the Game Master passes the player-generated code to the virtual simulator for execution, and if it is clarification or acknowledgement, it passes the response to the Programmer. If there are any errors in the code execution, the error feedback is passed to the Robot. The Robot may either correct the code or chose to ask for clarification from the Programmer if there are any ambiguities.

\subsubsection{Virtual Simulator}
\label{subsubsec:supp-vs}
The virtual simulator is used to execute the Robot-generated code to update the placement of components in the grid-based environment. As mentioned in Section~\ref{sec:methodology} of the main content, the grid-based environment uses a 2.5D grid of size 8x8, where each grid cell allows component placement and stacking. If the code execution is successful, the resultant world is returned with cell-by-cell stacking details to the Game Master, which are then added to the Programmer's prompt in the next turn. If there are failures during execution, the error information is passed to the Robot for further processing.

\subsection{Prompt Structure}
\label{subsec:appendix-prompts}
We follow the standard~\citep{DBLP:conf/nips/BrownMRSKDNSSAA20, DBLP:conf/nips/Wei0SBIXCLZ22, DBLP:journals/csur/LiuYFJHN23} prompting approaches to structure our Programmer and Cobot prompts. Both players are prompted in zero-shot settings. Each prompt consists of a system message, task description, environment information along with the constraints, followed by the expected response structure and the dialogue history up to that turn. Additionally, Programmer's prompt consists the difference board (i.e., the cell-level differences between the goal board and the Robot's reconstructed board) which can guide the Programmer in generating the follow-up instruction. In order to ensure consistent boundaries for the model inputs and outputs, we added response anchors (inspired by DSPy~\citep{DBLP:journals/corr/abs-2310-03714}) [[ \#\# prompt \#\# ]] and [[ \#\# completed \#\# ]] at the beginning and end of each prompt.  The interaction will be aborted if any of the players' responses do not follow the response constraints and will continue until the Programmer responds with \texttt{DONE} or the maximum number of turns ($15$) is reached. More details about the prompt ablation studies are available in the supplementary material.



\subsection{Prompt Ablation Studies}
\label{subsec:supp-prompts-ablation}
We conducted multiple experiments to refine the prompt templates and response constraints. We began by varying the number of turns required for the interaction from $1$ to $20$ and, based on the results (see Table~\ref{tab:suppturnsablation}), fixed it at $15$ turns. We then continued the ablations for the number of optimization turns and retry turns and chose $10$ and $3$, respectively, based on the results (see Tables~\ref{tab:suppoptimturnsablation} and~\ref{tab:suppretryturnsablation}). The next set of ablations focused on the necessity of including the difference grid, error feedback, and clarification questions. Table~\ref{tab:supprestablation} shows that the success rate decreased when these were not included in the prompt, and hence we retained them in the final prompt structure. Figures~\ref{fig:prompt_programmer_p1}-to-\ref{fig:prompt_robot_repeat} shows the prompt structure used across the experiments for both the players.

\begin{table*}[h]
\centering
\setlength{\tabcolsep}{10pt}
\caption{Overview of ablation results for the optimal number of turns required for the player interaction; A higher success rate indicates better performance.}
\label{tab:suppturnsablation}
\vspace{6pt}
\begin{tabular}{cccc}
\hline
Ablations & Abort Rate & Failure Rate & Success Rate \\ \hline
\textit{n}=1   & 0.013      & 0.642        & 0.345        \\
\textit{n}=3   & 0.006      & 0.558        & 0.436        \\
\textit{n}=5   & 0.006      & 0.448        & 0.545        \\
\textit{n}=10  & 0.012      & 0.424        & 0.564        \\
\textit{n}=15  & 0.0        & 0.382        & 0.618        \\
\textit{n}=20  & 0.0        & 0.380        & 0.620        \\ \hline
\end{tabular}%
\end{table*}

\begin{table*}[h]
\centering
\setlength{\tabcolsep}{10pt}
\caption{Overview of ablation results for the optimal number of turns required for the abstraction; A higher success rate indicates better performance.}
\label{tab:suppoptimturnsablation}
\vspace{6pt}
\begin{tabular}{cccc}
\hline
Ablations & Abort Rate & Failure Rate & Success Rate \\ \hline
\textit{n}=1   & 0.006  & 0.552 & 0.442        \\
\textit{n}=3   & 0.018  & 0.352 & 0.630        \\
\textit{n}=5   & 0.000  & 0.376 & 0.624        \\ \hline
\end{tabular}%
\end{table*}

\begin{table*}[h]
\centering
\setlength{\tabcolsep}{10pt}
\caption{Overview of ablation results for the optimal number of turns required for the Robot to retry in case of execution errors; A higher success rate indicates better performance.}
\label{tab:suppretryturnsablation}
\vspace{6pt}
\begin{tabular}{cccc}
\hline
Ablations & Abort Rate & Failure Rate & Success Rate \\ \hline
\textit{n}=1   & 0.182  & 0.224 & 0.594        \\
\textit{n}=3   & 0.039  & 0.276 & 0.685        \\
\textit{n}=5   & 0.000  & 0.309 & 0.691        \\ \hline
\end{tabular}%
\end{table*}

\begin{table*}[h]
\centering
\setlength{\tabcolsep}{10pt}
\caption{Overview of ablation results for including various types of feedback in the interaction; A higher success rate indicates better performance.}
\label{tab:supprestablation}
\vspace{6pt}
\begin{tabular}{cccc}
\hline
Ablations & Abort Rate & Failure Rate & Success Rate \\ \hline
No difference grid   & 0.182  & 0.224 & 0.594        \\
No error feedback   & 0.594  & 0.055 & 0.352        \\
No clarification questions   & 0.455  & 0.097 & 0.448        \\ \hline
\end{tabular}%
\end{table*}

\begin{figure*}[h]
  \centering
    \caption{Part-1 of prompt template used for the Programmer in the Teach step.}
    \label{fig:prompt_programmer_p1}  
    \begin{prompt}
\\
Your input fields are:

1. `prompt` (str): Prompt with grid\_size, skillname, goal grid, difference\_grid
\\\\
Your output fields are:

1. `instruction` (str): Natural-language instruction for the robot

All interactions will be structured in the following way, with the appropriate values filled in.
\\\\

[[ \#\# prompt \#\# ]]

{prompt}

[[ \#\# instruction \#\# ]]

{instruction}

[[ \#\# completed \#\# ]]
\\\\
In adhering to this structure, your objective is: 

        You are the human user.

        Your task is to collaborate with a robot to reconstruct the given goal grid.
\\\\
        Preferred Instruction Style:

        • Describe shapes, regions, and patterns.

        • Use relative spatial language (center, corners, edges, top/bottom/left/right).

        • Use symmetry, repetition, and grouping where possible.

        • Describe what the grid should look like, and how to reconstruct it
\\\\
        Communication Strategy:

        • Focus on completing one meaningful pattern or region at a time.

        • Assume the follower understands shapes and patterns.   
\\\\
        Stacking constraints:

        • Instructions must respect stacking dependencies.

        • Always describe actions from bottom to top: supporting objects must be placed before any objects that rest on them.

        • Do NOT instruct placement of top-level objects before their supports exist.

        • Violating this order may lead to incorrect reconstruction.
\\\\
        Rules:

        • Use the difference\_grid to identify which high-level patterns or regions are incomplete.

        • If a region or pattern is already correct, do not mention it.

        • Generates the instructions to build incorrect or missing elements.

        • If clarification input is non-empty, address it directly and clearly.

        • If the entire grid matches the goal grid and reconstruction\_status is True, reply only with the word "DONE".

        • Otherwise, generate instructions that advances the reconstruction

        • Always output either an instruction or "DONE" — nothing else.
      
\end{prompt}

\end{figure*}        

\begin{figure*}[h]
  \centering
    \caption{Part-2 of prompt template used for the Programmer in the Teach step.}
    \label{fig:prompt_programmer_p2}  
    \begin{prompt}
\\
        Examples:
        
        "Create a vertical stack with a red washer below a blue screw."

        "Attach a yellow nut to the right side of green horizontal bridge."

        "Mirror the structure on the left side onto the right side."

        "First place the nut, then place screw on top of it."
\\\\
Inputs:

    - prompt: Prompt with the following details:

        - grid\_size: The grid is of size {grid\_size}.

        - goal\_grid: The goal configuration of the grid            

        - difference\_grid: The difference between the grid filled by the robot and goal grid

        - clarification: Optional question received from the robot; may be empty.

        - acknowledgment: Optional message received from the robot; may be empty.

        - reconstruction\_status: True if the current grid matches the goal grid, False otherwise.\\\\

Respond with the corresponding output fields, starting with the field `[[ \#\# instruction \#\# ]]`, and then ending with the marker for `[[ \#\# completed \#\# ]]`.
\end{prompt}

\end{figure*}        
\begin{figure*}[h]
  \centering
    \caption{An example goal board representation.}
    \label{fig:prompt_goal_board}  
    \begin{prompt}
\\
Grid size: 8x8

Object name: wn

Colors: ['red', 'green']

Location: [3, 1]
\\\\
Goal:

Grid levels (bottom to top):

Level 1:

row: 3, col: 1: {'shapes': ['washer'], 'colors': ['red']}

Level 2:

row: 3, col: 1: {'shapes': ['nut'], 'colors': ['green']}

\end{prompt}

\end{figure*}   

\begin{figure*}[h]
  \centering
    \caption{An example difference board representation.}
    \label{fig:prompt_difference_board}  
    \begin{prompt}
\\
Goal:Grid levels (bottom to top):

Level 1:

row: 3, col: 1: {`shapes': [`washer'], `colors': [`red']}

Level 2:

row: 3, col: 1: {`shapes': [`nut'], `colors': [`green']}
\\\\
Difference grid: Differences between target grid and player's grid (bottom to top):

Level 1:

- Expected: {'shapes': ['washer'], 'colors': ['red']}

- Current Status: {'shapes': ['screw'], 'colors': ['red']}

Level 2:

row: 1, col: 3:

- Missing: {'shapes': ['nut'], 'colors': ['green']}

\end{prompt}

\end{figure*}   
\begin{figure*}[h]
  \centering
    \caption{Part-1 of prompt template used for the Robot in the Teach step.}
    \label{fig:prompt_robot_p1}  
    \begin{prompt}
\\

Your input fields are:

1. `prompt` (str): Prompt with Instruction from User and current grid state
\\\\
Your output fields are:

1. `player\_response` (str): A JSON object with keys 'status' and 'details' as described above
\\\\

All interactions will be structured in the following way, with the appropriate values filled in.

[[\#\# prompt \#\# ]]
{prompt}

[[ \#\# player\_response \#\# ]]

{player\_response}

[[ \#\# completed \#\# ]]
\\\\
In adhering to this structure, your objective is: 

        You are the robot (the instruction follower) in a collaborative grid-filling game.
\\\\
        Interpretation Mode:

        • Instructions may describe patterns, regions, relationships, or rules rather than exact cells.

        • Ground high-level instructions using shared spatial and stacking conventions.
\\\\
        Spatial Conventions:

        • "to the right of X" = horizontally adjacent in the same row

        • "to the left of X" = horizontally adjacent in the same row

        • "above X" / "below X" = vertically adjacent in the same column

        • "stacked on X" = same cell, higher level

        • "attached to X" = adjacent, not stacked

        • If a spatial relation is unclear, ask for clarification. 
\\\\
        Ordering Constraints (MANDATORY):
        • Actions must respect stacking dependencies.

        • Always place supporting (lower-level) shapes before shapes that rest on them.

        • Never place a top-level shape before its support exists.

        • If an instruction implies an invalid order, ask for the clarification

        • Assume instructions are executed immediately; do not anticipate future steps.

        Examples:

        Instruction:

        "Place screws on top of washers"
\\\\
        Correct:

            First place washers, then place screws on top of them
\\\\
        InCorrect (order violation):

            screws placed before washers exist
\end{prompt}
\end{figure*}        

\begin{figure*}[h]
  \centering
   \caption{Part-2 of prompt template used for the Robot in the Teach step.}
    \label{fig:prompt_robot_p2}  
    \begin{prompt}
\\
        Input Instruction Examples:

        You receive an instruction such as:

        "Place a washer with a nut stacked on top." or

        "Attach a yellow nut to the right of each green horizontal bridge."

        • If coordinates are explicitly given, use them directly.

        • If instructions are high-level (patterns, regions, relations), infer coordinates using the current grid state and shared spatial conventions.

        • If inference is ambiguous, request clarification instead of guessing.
\\\\
        Clarification policy:

        • If an instruction is underspecified (location, count, or relation unclear) or ambiguous, respond with status="clarification" before executing.

        • Do NOT invent coordinates, colors, types or quantities.

        • Request clarification only when the required information is not recoverable from the current grid state or prior instructions.

        • If an instruction contradicts stacking or physical constraints, request clarification instead of repairing it.

        Examples:

          Instruction:

            "Place screw under the nut"

          Correct:
            Ask for clarification because the instruction violates stacking order.

            Example clarification: "Placing a screw under a nut would require placing the nut first, which violates the bottom-to-top rule. 

                                   Did you intend the screw to be stacked on top of the nut?"

          InCorrect (order violation): DO NOT DO THIS

            Placing the nut first and then stacking a screw under it
            Silently reversing the order without asking       
\\\\
          Instruction:

            "Stack a horizontal bridge"

          Correct:

            If the current grid does NOT satisfy the depth requirements for placing a horizontal bridge at the intended span, ask for clarification.

            Example clarification: "I don’t see any span where the depth requirements for a horizontal bridge are met. Can you clarify the intended placement?" 

          InCorrect (depth violation): DO NOT DO THIS

            Choosing a random shape to stack the bridge on
            Assuming default depth support and placing the bridge
\\\\
\end{prompt}
\end{figure*}        

\begin{figure*}[h]
  \centering
    \caption{Part-3 of prompt template used for the Robot in the Teach step.}
    \label{fig:prompt_robot_p3}  
    \begin{prompt}    
        Available APIs:

        - put(board: np.ndarray, shape: str, color: str, x: int, y: int)
            • Places a shape of a given color on the board at coordinates (x, y).

            • Coordinates may be provided explicitly OR implied by the instruction + current\_grid.

                - If the instruction explicitly specifies row/column/cell indices, use them directly.

                - If the instruction specifies a relative location (e.g., "to the right of X", "above X",

                  "in the top-left corner", "center", "top row", "leftmost column"), infer (x, y) from instruction and current\_grid
                  using the spatial conventions.

                - If multiple (x, y) satisfy the instruction or the intended location is unclear, DO NOT guess. Ask for clarification.

            • Examples of inference:

                - "to the right of the green washer at (r,c)" → (x,y) = (r, c+1)

                - "top-left corner" → (x,y) = (1,1)

                - "top row, third cell" → infer as (x=top\_row (1), y=3)

                - "bottom-right corner" → (x,y) = (8,8)

        - move(board: np.ndarray, x1: int, y1: int, x2: int, y2:int, shapes\_list: List = None)
            
            • To move a shape (or stack of shapes) from coordinates (x1, y1) to coordinates (x2, y2) on the board

            • Coordinates may be provided explicitly OR implied by the instruction + current\_grid.

                - If the instruction explicitly specifies row/column/cell indices, use them directly.

                - If the instruction specifies a relative location (e.g., "to the right of X", "above X",
                  "in the top-left corner", "center", "top row", "leftmost column"), infer (x, y) from instruction and current\_grid
                  using the spatial conventions.

                - If multiple (x, y) satisfy the instruction or the intended location is unclear, DO NOT guess. Ask for clarification.

            • Examples of inference:

                - "to the right of the green washer at (r,c)" → (x,y) = (r, c+1)

                - "top-left corner" → (x,y) = (1,1)

                - "top row, third cell" → infer as (x=top\_row (1), y=3)

                - "bottom-right corner" → (x,y) = (8,8)

            • Do not use same coordinates for source and destination

            • Do not use this function to remove shapes from the board, use removeshape instead

            • Use x1, y1 as source coordinates and x2, y2 as destination coordinates

            • Use x1, y1, x2 and y2 do not use same names for source and destination

            • x1, x2: Row values

            • y1, y2: Column values

            • Examples:

              move it from 1st column to last column in the 5th row then x1=5, y1=1, x2=5, y2=8

              move it from 3rd to next row, then x1=3, x2=4

              move it from 7th column to 8th column, then y1=7, y2=8

              move it to 8th column, then y2=8

            • shape\_list: (List, optional): List of shapes from bottom → top to move. If None, only the topmost shape in the stack is moved.
\end{prompt}
\end{figure*}        

\begin{figure*}[h]
  \centering
    \caption{Part-4 of prompt template used for the Robot in the Teach step.}
    \label{fig:prompt_robot_p4}  
    \begin{prompt}
\\        
        - removeshape(board: np.ndarray, x: int, y: int, shape, color)

            • To remove a shape from coordinates (x1, y1) on the board

            • Coordinates may be provided explicitly OR implied by the instruction + current\_grid.

                - If the instruction explicitly specifies row/column/cell indices, use them directly.

                - If the instruction specifies a relative location (e.g., "to the right of X", "above X",
                  "in the top-left corner", "center", "top row", "leftmost column"), infer (x, y) from instruction and current\_grid
                  using the spatial conventions.

                - If multiple (x, y) satisfy the instruction or the intended location is unclear, DO NOT guess. Ask for clarification.

            • Examples of inference:

                - "to the right of the green washer at (r,c)" → (x,y) = (r, c+1)

                - "top-left corner" → (x,y) = (1,1)

                - "top row, third cell" → infer as (x=top\_row (1), y=3)

                - "bottom-right corner" → (x,y) = (8,8)

            • Do not use this function to move shapes, use move instead

            • Use this function only when there is a placement error. If a shape already exists in the target cell, ask whether it should be removed before placing a new one. In most cases, you will be stacking a new shape in the same cell, which doesn’t require removeShape();just call put() for the new shape.

            • The shape and color must match the topmost shape in the stack at (x, y)                
            • x: Row  (If the human says "row = 3", then use x = 3)

            • y: Column  (If the human says "column = 1", then use y = 1)

            • If human says remove it from 6th row, then x=6

            • If human says remove it from 1st column, then y=1

            • shape\_list: (List, optional): List of shapes from bottom → top to move. If None, only the topmost shape in the stack is moved.

        - clear(board: np.ndarray)

            • To clear all the shapes on the board

        - undo(board: np.ndarray)

            • To undo last placement action(s) on the board

            • If multiple shapes were placed as part of the previous operation, all of them will be cleared from the board.            
        
        You receive an instruction such as:
            "Place a washer and nut (green and blue) in cell (1, 8)."
\\\\        
        ---
        AVAILABLE SHAPES

          - washer

          - nut

          - screw

          - bridge (vertical, horizontal)  ← special shape

             • Use **'bridge-h'** for horizontal bridges and **'bridge-v'** for vertical ones.      
\\\\
      
\end{prompt}
\end{figure*}        

\begin{figure*}[h]
  \centering
    \caption{Part-5 of prompt template used for the Robot in the Teach step.}
    \label{fig:prompt_robot_p5}  
    \begin{prompt}
\\                
        AVAILABLE COLORS

          - green

          - red

          - blue

          - yellow
\\\\         
        SHAPE OCCUPANCY

          • All shapes occupy **exactly one cell**, **except the "bridge"**.

          • A **bridge** spans **two adjacent cells**:
              
              - *Horizontal bridge*: spans consecutive **columns** in the same row.
              
              - *Vertical bridge*: spans consecutive **rows** in the same column.

          • A bridge **requires two other shapes** underneath it (referred as depth) for support while stacking, one under each end of the bridge.

        BRIDGE REPRESENTATION RULES

          • Bridges span two cells and appear in the target grid as:

              - bridge-v-top / bridge-v-bottom  → one vertical bridge
              
              - bridge-h-left / bridge-h-right  → one horizontal bridge
              
              - 'bridge-h-left' and 'bridge-h-right' are display labels representing the two halves of a single horizontal bridge (bridge-h). A horizontal bridge always occupies two adjacent cells in the same row. When generating code or interpreting grid , always refer to the shape as bridge-h, not the left/right variants.
              
              - Similarly 'bridge-v-top' and 'bridge-v-bottom' are display labels representing the two halves of a single vertical bridge 'bridge-v'. A vertical bridge always occupies two adjacent cells in the same column. When generating code or interpreting grid , always refer to the shape as bridge-v, not the top/bottom variants.

          • Example:
              
              Instruction: Place a green horizintal bridge at the bottom left corner
              
              → Code: put(board, 'bridge-h', 'green', 8, 1)
              
              → Current-Grid would become:
              
                Grid levels (bottom to top):
                
                Level 1:
                
                row: 8, col: 1: {'shapes': ['bridge-h-left'], 'colors': ['green']}
                
                row: 8, col: 2: {'shapes': ['bridge-h-right'], 'colors': ['green']}
              
              → To remove this shape: removeshape(board, x=8, y=1, shape='bridge-h', color='green')

              Instruction: Stack a blue vertical bridge at (2,8)
              
              → Code: put(board, 'bridge-v', 'blue', 2, 8)
              
              → Current-Grid would become:
              
                Grid levels (bottom to top):
                
                Level 1:
                
                row: 2, col: 8: {'shapes': ['bridge-v-top'], 'colors': ['blue']}
                
                row: 3, col: 8: {'shapes': ['bridge-v-bottom'], 'colors': ['blue']} 
              
              → To remove this shape: removeshape(board, x=2, y=8, shape='bridge-v', color='blue')                                              
\\\\
\end{prompt}
\end{figure*}        

\begin{figure*}[h]
  \centering
    \caption{Part-6 of prompt template used for the Robot in the Teach step.}
    \label{fig:prompt_robot_p6}  
    \begin{prompt}
\\          
        STACKING \& DEPTH RULES
        
          • Shapes can be stacked vertically within the same cell.
          
          • Stacking is only allowed if all shapes share the **same depth**.
          
          • Since the shapes can be stacked, multiple shapes may occupy the same cell. If you have any questions about stacking different shapes in one cell, feel free to ask for clarification. You do not need to remove an existing shape to stack a new one in the same cell.            
\\\\            
            Example:
            
            Instruction: Place a red washer in top left corner
            
                Correct: put(board, shape='washer', color='red', x=1, y=1)
\\\\            
            Instruction: Stack a green nut in  top left corner
                
                Correct: put(board, shape='nut', color='green', x=1, y=1)

                Incorrect: put(board, shape='washer', color='red', x=1, y=1)
                           removeshape(board, x=1, y=1, shape='washer', color='red')
                           put(board, shape='nut', color='green', x=1, y=1)
\\\\            
            Instruction: Place a blue horizontal bridge in the middle of the grid
            
                Correct: put(board, shape='bridge-h', color='blue', x=4, y=4)
\\\\            
            Instruction: Stack the bridge with a yellow screw on the left
            
                Correct: put(board, shape='screw', color='yellow', x=4, y=4)

                Incorrect: put(board, shape='bridge-h', color='blue', x=4, y=4)
                           removeshape(board, x=4, y=4, shape='bridge-h', color='blue')
                           put(board, shape='screw', color='yellow', x=4, y=4)   
\\\\
            Instruction: Place a yellow vertical bridge in bottom right corner
                
                Correct: put(board, shape='bridge-v', color='yellow', x=7, y=8)
\\\\            
            Instruction: Stack a green washer in the bottom
            
                Correct: put(board, shape='washer', color='green', x=8, y=8)

                Incorrect: put(board, shape='bridge-v', color='yellow', x=7, y=8)
                           removeshape(board, x=7, y=8, shape='bridge-v', color='yellow')
                           put(board, shape='washer', color='green', x=8, y=8)                                                     
\\\\
          • When multiple shapes are placed in a single cell:
              
              - The **first mentioned shape** is placed **at the bottom**.
              
              - Later shapes stack **on top**.
          
          • Do not stack same shapes in the same cell.
          
            Example:
\\\\            
                Correct: put(board, shape='washer', color='red', x=1, y=1)
                
                         put(board, shape='nut', color='green', x=1, y=1)
\\\\                
                Incorrect: put(board, shape='washer', color='red', x=1, y=1)
                         
                           put(board, shape='washer', color='green', x=1, y=1)                
\end{prompt}

\end{figure*}        

\begin{figure*}[h]
  \centering
    \caption{Part-7 of prompt template used for the Robot in the Teach step.}
    \label{fig:prompt_robot_p7}  
    \begin{prompt}
\\ 
        ---

        IMPLEMENTATION DETAILS

          • When placing **multiple shapes**, use **loops** or repeated `put` calls as needed.

            Example:

                for row in [1,4]:
                    
                    put(board, shape, color, row, y)
        
                for col in [2,5]:
                    
                    put(board, shape, color, x, col)
        
                for row in [1,3]:
                    
                    for col in [1,4]:
                    
                        put(board, shape, color, row, col)
        
          • **Do not assume** any unspecified details (e.g., color, orientation, or position).
            
            If an instruction is ambiguous, **ask Player 1 for clarification** instead of guessing.
        
        --- 

        ---

        CURRENT GRID

          • The current grid lists the current status of reconstruction
          
          • Each cell may be marked as one of the shapes: 'washer', 'nut', 'screw', 'bridge-h-left', 'bridge-h-right', 'bridge-v-top', and 'bridge-h-bottom'
          
            - 'bridge-h-left' and 'bridge-h-right' are display labels representing the two halves of a single horizontal bridge (bridge-h). A horizontal bridge always occupies two adjacent cells in the same row. When interpreting instructions/current grid/interpreting actions (put / remove / move), always refer to the shape as bridge-h, not the left/right variants.
            
            - Similarly 'bridge-v-top' and 'bridge-v-bottom' are display labels representing the two halves of a single vertical bridge 'bridge-v'. A vertical bridge always occupies two adjacent cells in the same column. When interpreting instructions/current grid/interpreting actions (put / remove / move), always refer to the shape as bridge-v, not the top/bottom variants.
          
          • **Do not assume** any unspecified details (e.g., color, orientation, or position).
          
            If an instruction is ambiguous, **ask Player 1 for clarification** instead of guessing.
        
        ---    
\\\\
        Given:
        
        - prompt: Prompt with the following details:
        
            - grid\_size: The grid is of size {grid\_size}.
            
            - The instruction from the user describing what to do
            
            - current\_grid: The current filled cells of the grid, None if empty

\end{prompt}
\end{figure*}        

\begin{figure*}[h]
  \centering
    \caption{Part-8 of prompt template used for the Robot in the Teach step.}
    \label{fig:prompt_robot_p8}  
    \begin{prompt}
\\ 
        Produce:

        - Respond with a JSON object:
          
          Format:
          
          {
          
            "status": "<string>",       \# e.g. "clarification" or "code"
            
            "details": "<string>",     \# Python code when status="code", or plain text when 
            
            status="clarification"
          
          }

          If clarification needed:
          
          {
          
            "status": "clarification",
            
            "details": "Which color washer should I use?"
          
          }

          If providing acknowledgement:
          
          {
          
            "status": "acknowledgement",
            
            "details": "Yes, I know"
         
          }

          If executing code:
          
          {
          
               "status": "code",

               "details": "put(board, 'washer', 'green', x=1, y=1)
               
               put(board, 'nut', 'blue', x=3, y=2)"
          
          }

        Make sure the JSON is valid and parsable by Python json.loads().
Respond with the corresponding output fields, starting with the field `[[ \#\# player\_response \#\# ]]`, and then ending with the marker for `[[ \#\# completed \#\# ]]`.

\end{prompt}

\end{figure*}        

\begin{figure*}[h]
  \centering
    \caption{Part-1 of prompt template used in the Abstract step.}
    \label{fig:prompt_abstract_p1}  
    \begin{prompt}
\\
Your input fields are:

1. `prompt` (str): Prompt with function name, usage, instruction-code pairs, and optional error feedback.

Your output fields are:\\\\
Your output fields are:

1. `optimized\_function` (str): A single optimized Python function as a string.

All interactions will be structured in the following way, with the appropriate values filled in.
\\\\

[[ \#\# prompt \#\# ]]

{prompt}

[[ \#\# optimized\_function \#\# ]]

{optimized\_function}

[[ \#\# completed \#\# ]]
\\\\
In adhering to this structure, your objective is: 

        Given multiple instruction–code pairs, generate a single reusable and optimized
        Python function that performs all described operations in one call.
\\\\
        The optimized function should:
        
        - Use the provided function definition exactly as given.
        
        - Always use board as first argument to the function and followed by colors (a list of colors) and x, y (the starting/reference point row and column).
        
        - Do not use shapes as parameters. Keep shapes as fixed attributes (e.g., shapes like 'washer', 'nut', 'bridge') and use them only inside the function. Use the  exact shape names as in the code snippets.
        
        - Do not hardcode color values inside the function.
        
        - Use lists for parameters when multiple values of the same type are needed (e.g., multiple colors).
        
        - All put() operations are relative to the starting position (x, y) specified in the first instruction. Use that as the reference point and as an input parameter to the function. The function should compute the relative offsets of all other shapes based on how their coordinates differ from the first placement.
        
           - If the first instruction places a shape at (2, 3) and another places one at (2, 4), then the function input should be (x=2, y=3).
           
           - The second shape’s position is relative to the first by an offset of (dx=0, dy=1), which should be computed and applied inside the function.
        
        - After determining the relative offsets for each placement, factor out repetition.
        
        - Use a concise loop to iterate through corresponding shapes, colors, and offsets, rather than writing separate put() statements for each shape.
        
        - Do not change the order of operations
        
        - Do not change the signature of the function in the input code
          
          - Here is the signature of put(): put(board, shape, color, x, y)
        
        - Wherever multiple put() calls can be combined use loops. Do not repeat similar lines of code.
\end{prompt}
\end{figure*}        

\begin{figure*}[h]
  \centering
    \caption{Part-2 of prompt template used in the Abstract step.}
    \label{fig:prompt_abstract_p2}  
    \begin{prompt}
\\        
        
        - When executed once, it should perform all of the described actions together.

        - Use clean, valid Python syntax (PEP-8 compliant).
        
        - Maintain semantic equivalence with all provided code snippets.
        
        - Do not reuse the input code–sequence pairs as they are; the sequence is meant to be optimized.
        
        - Output a valid standalone Python optimized function.
\\\\
        API Interpretation

        - The input code snippets contains some or all of these APIs.
              - put(board: np.ndarray, shape: str, color: str, x: int, y: int): places a shape on the board at (x1,y1)
              - move(board: np.ndarray, x1: int, y1: int, x2: int, y2:int, shapes\_list: List = None): moves a shape from (x1,y1) to (x2,y2)
              - removeshape(board: np.ndarray, x: int, y: int, shape, color): removes a shape from (x1,y1)
              - clear(board: np.ndarray): clears/removes all shapes from the board
              - undo(board: np.ndarray): reverses the last action(s) 
\\\\
        - You must treat the snippet as a sequence of actions executed on an initially empty board.
              
              - Step through the actions in order and reason about how each one changes the board.
              
              - Actions like remove(), move(), undo(), and clear() can cancel or override earlier put() calls.
              
              - Any shape that is removed, moved away, undone, or cleared should NOT appear in the final board state. Earlier put() calls that get canceled in this way should be ignored.
              
              - Figure out which shapes actually remain at the end.
\\\\

        - Your goal is to figure out what shapes are on which cells *after the entire sequence has finished*.
             
             - For example:
             
               put(board, 'washer', 'red', 1, 1)
               
               remove(board, 1, 1, 'washer', 'red')
               
               put(board, 'nut', 'blue', 1, 1)
               
               At the end, only a blue 'nut' remains at (1, 1). The red 'washer' should NOT be included in the final result.

               put(board, 'bridge-h', 'green', 2, 7)
               
               put(board, 'screw', 'yellow', 6, 3)
               
               move(board, 6, 3, 8, 4, ['screw'])
               
               At the end, bridge-h is at (2,7) and screw is at (8,4). There is no screw at (6,3)
\\\\
        - After you have inferred the final board state, generate an optimized function that:
             - Uses ONLY put() calls
             
             - Do NOT use move(), remove(), undo(), or clear() in the optimized function.
\end{prompt}
\end{figure*}        

\begin{figure*}[h]
  \centering
    \caption{Part-3 of prompt template used in the Abstract step.}
    \label{fig:prompt_abstract_p3}  
    \begin{prompt}
\\  
        SHAPE OCCUPANCY
        - All shapes occupy **exactly one cell**, **except the "bridge"**.
        
        - A **bridge** spans **two adjacent cells**:
        
              - *Horizontal bridge*: spans consecutive **columns** in the same row.
              
              - put(board, 'bridge-h', 'red', x=0,y=0): This places bridge-h in cells (0,0) and (0,1): Same row, consequent columns
              
              - *Vertical bridge*: spans consecutive **rows** in the same column.
              
              - put(board, 'bridge-v', 'red', x=0,y=0): This places bridge-v in cells (0,0) and (1,0): Consequent rows, same columns              
        
        - A bridge **requires two other shapes** underneath it for support while stacking,
            one under each end of the bridge.
\\\\

        STACKING \& DEPTH RULES
        
        - Shapes can be stacked vertically within the same cell.
        
        - Stacking is only allowed if all shapes share the **same depth**.
        
        - When multiple shapes are placed in a single cell:
        
              - The **first mentioned shape** is placed **at the bottom**.
              
              - Later shapes stack **on top**.
        - Do not stack same shapes in the same cell.
\\\\            
            Example:
                
                Correct: 

                         put(board, shape='washer', color='red', x=0, y=0)
                
                         put(board, shape='nut', color='green', x=0, y=0)
\\\\                
                Incorrect: 

                           put(board, shape='washer', color='red', x=0, y=0)
                        
                           put(board, shape='washer', color='green', x=0, y=0)    
\\\\
        
        Given:
        
        - prompt: Prompt with the following details:
            
            - function signature: signature of the function including its name. use only this signature.
            
            - function usage: how the function will be used
            
            - instruction-code snippets across turns: a list of code snippets corresponding to different instructions
            
            - optional error feedback from previous attempts
        
        Produce:
            
            - A single optimized Python function as a string.
            
            - Do not generate any other reasoning traces as it will fail code execution
        
        Make sure the Python function is valid and executable by executing `exec()` on it.
\\\\
Respond with the corresponding output fields, starting with the field `[[ \#\# optimized\_function \#\# ]]`, and then ending with the marker for `[[ \#\# completed \#\# ]]`.
\end{prompt}
\end{figure*}        
\begin{figure*}[h]
  \centering
    \caption{Prompt template used for the Programmer in the Reuse step.}
    \label{fig:prompt_programmer_reuse_p1}  
    \begin{prompt}
\\
Your input fields are:

1. `prompt` (str): Prompt with grid\_size, skillname, goal grid, difference\_grid
\\\\
Your output fields are:

1. `instruction` (str): Natural-language instruction for the robot

All interactions will be structured in the following way, with the appropriate values filled in.
\\\\

[[ \#\# prompt \#\# ]]

{prompt}

[[ \#\# instruction \#\# ]]

{instruction}

[[ \#\# completed \#\# ]]
\\\\
In adhering to this structure, your objective is: 

        You are the human user.

        Your task is to collaborate with a robot to reconstruct the given goal grid
\\\\
        You are reconstructing an object that was previously reconstructed by robot with different attributes.  The object’s internal structure is unchanged. Confirm whether the robot knows how to reconstruct the object. If so, specify only what differs (e.g., location, colors). If the robot indicates it doesn't know how to reconstruct the object, you may then explain the reconstruction steps. The structure of the object to be described explicitly is mentioned under Layerwise Representation. **DO NOT** explain explicit placement steps unless the robot indicates that it does not know how to reconstruct the object.
\\\\
        The rest of the prompt (from Preferred Instruction Style to Rules) is the same as in the Teach step.\\\\

Decision Protocol:

        - Assume that the robot is aware of reconstructing the object using the existing skills. → Output an instruction to reconstruct <object\_name> specifying the exact color values and placement location necessary to build the target object. You **MUST NOT** explain shape-wise/layer-wise placement steps.

        - If the robot is not aware (e.g., the robot responds with "No, I don’t know about the xxx object/skill") → Output: The reconstruction process layer by layer (from bottom to top), by including the colors and location information. **DO** this only when robot has explicitly confirmed it does not know how to reconstruct the object.
\\\\
The rest of the prompt (from Inputs onwards) is the same as in the Teach step.
\end{prompt}

\end{figure*}        
\begin{figure*}[h]
  \centering
    \caption{Prompt template used for the Robot in the Reuse step.}
    \label{fig:prompt_robot_p1}  
    \begin{prompt}
\\

Your input fields are:

1. `prompt` (str): Prompt with Instruction from User and current grid state
\\\\
Your output fields are:

1. `player\_response` (str): A JSON object with keys 'status' and 'details' as described above
\\\\

All interactions will be structured in the following way, with the appropriate values filled in.

[[\#\# prompt \#\# ]]
{prompt}

[[ \#\# player\_response \#\# ]]

{player\_response}

[[ \#\# completed \#\# ]]
\\\\
Your updated objective is:

        You are reconstructing an object that was previously built, with some properties (e.g., location or colors) changed.

        If a relevant skill is available (under LEARNED SKILLS), you **MUST** use it. If no suitable skill exists, acknowledge the same to user and follow the user instructions.        
\\\\
      The rest of the prompt (from Interpretation Mode onwards until Correction) is the same as in the Teach step.\\\\
      
      LEARNED SKILLS
      
      \$LEARNEDSKILLS\\\\

      The rest of the prompt (from Available APIs onwards) is the same as in the Teach step.

\end{prompt}

\end{figure*}        

\begin{figure*}[h]
  \centering
    \caption{Prompt template used for the Programmer in the Repeat step.}
    \label{fig:prompt_programmer_repeat}  
    \begin{prompt}
\\
Your input fields are:

1. `prompt` (str): Prompt with grid\_size, skillname, goal grid, difference\_grid
\\
Your output fields are:

1. `instruction` (str): Natural-language instruction for the robot

All interactions will be structured in the following way, with the appropriate values filled in.
\\

[[ \#\# prompt \#\# ]]

{prompt}

[[ \#\# instruction \#\# ]]

{instruction}

[[ \#\# completed \#\# ]]
\\\
In adhering to this structure, your objective is: 

        You are the human user. Your task is to collaborate with a robot to reconstruct the given goal grid.
\\
        The rest of the prompt (from Preferred Instruction Style to Rules) is the same as in the Teach step.\\\\

Decision Protocol:

        - Assume that the robot is aware of reconstructing the object using the existing skills. → Output an instruction to reconstruct <object\_name> specifying the exact color values and placement location necessary to build the target object. You **MUST NOT** explain shape-wise/layer-wise placement steps.

        - When generating instructions to describe the pattern, list the colors in the same order as provided under Colors in the prompt. The order must follow the structure of the object: all bottom-layer cells first, followed by the cells in the upper layers. Do not change this order.   

        - If the robot is not aware (e.g., the robot responds with "No, I don’t know about the xxx object/skill") → Output: The reconstruction process layer by layer (from bottom to top, not cell by cell), by including the colors and location information. **DO** this only when robot has explicitly confirmed it does not know how to reconstruct the object.

        - When robot has explicitly confirmed it does not know how to reconstruct the object, explain the object layer by layer, proceeding from bottom to top. When multiple layers are present, describe all relevant shapes in the bottom layer first across the full structure before moving to the next layer. Do not describe on cell basis from bottom to top before describing the other bottom cells, as this can lead to mistakes and misinterpretation.

        - When answering clarification questions, ensure that the robot’s concern is aligned with the goal specification. Verify whether the locations or colors identified as conflicting actually match the goal. If not, correct the misunderstanding.

        - If the robot cannot resolve the error after multiple attempts, guide it to clear the current grid and restart by using the appropriate skill API.              
\\\\
The rest of the prompt (from Inputs onwards) is the same as in the Teach step.
\end{prompt}

\end{figure*}        
\begin{figure*}[h]
  \centering
    \caption{Prompt template used for the Robot in the Repeat step.}
    \label{fig:prompt_robot_repeat}  
    \begin{prompt}
\\
Your input fields are:

1. `prompt` (str): Prompt with grid\_size, skillname, goal grid, difference\_grid
\\\\
Your output fields are:

1. `instruction` (str): Natural-language instruction for the robot

All interactions will be structured in the following way, with the appropriate values filled in.
\\\\

[[ \#\# prompt \#\# ]]

{prompt}

[[ \#\# instruction \#\# ]]

{instruction}

[[ \#\# completed \#\# ]]
\\
In adhering to this structure, your objective is: 

        You are the human user.

        Your task is to collaborate with a robot to reconstruct the given goal grid
\\\\
        The rest of the prompt (until Current Grid) is the same as in the Reuse step.\\

        Error Handling and Code Generation Rules

          • If executing the generated code triggers a compilation error and resolving it requires additional information, ask the user for clarification.

          • Do not repeatedly generate the same code across multiple turns. If the error message indicates the issue (e.g., placing a shape outside the grid), fix the problem by consulting with the user before generating new code.

        Grid State Awareness

          • The generated code runs on top of the current grid state.

          • Before generating new code, check what already exists in the grid and generate only the required changes.

          • Do not regenerate shapes that already exist unless modification is required.

        Sequential Execution Awareness

          • Each line of generated code modifies the grid state.

          • After line 1 executes, the grid is already updated.

          • Ensure that later lines do not conflict with changes introduced by earlier lines.

        Skill Name Accuracy and Usage

          • Skill names are sensitive. Example: wn, wns, and wb are different skills.

          • If a skill appears unavailable: Verify that the skill name matches the instruction. If the name is correct but still unavailable, ask the user how to build it. Do not continue using an incorrect skill.

          • Do not import any additional skill libraries. Use the skill API directly with the correct arguments. Ensure that the number of colors equals the number of shapes specified in the function or object description.
\\\\
The rest of the prompt (from Given onwards) is the same as in the Teach step.
\end{prompt}

\end{figure*}        

\end{document}